\documentclass[10pt,journal,compsoc]{IEEEtran}
\ifCLASSOPTIONcompsoc
  \usepackage[nocompress]{cite}
\else
  \usepackage{cite}
\fi
\ifCLASSINFOpdf
\else
\fi
\usepackage{booktabs} 
\usepackage{subfig}
\usepackage{graphicx}
\usepackage{algorithm}
\usepackage{algpseudocode}%
\usepackage{xcolor}
\usepackage{amsmath, amsthm}
\usepackage{amssymb}

\usepackage{mathtools}
\usepackage{multirow}
\usepackage{graphics}
\usepackage{enumitem}
\usepackage{CJKutf8}
\usepackage{array}
\usepackage{diagbox}
\usepackage{svg}
\usepackage{url}
\usepackage{threeparttable}
\usepackage{comment}
\usepackage{pifont}
\usepackage{tabularx}
\usepackage{cite,etoolbox}
\usepackage[dvipsnames]{xcolor}
\usepackage[table]{xcolor}
\definecolor{mypurple}{rgb}{0.4392, 0.1882, 0.6275}
\definecolor{darkblue}{rgb}{0.0, 0.0, 0.55}
\definecolor{darkgray}{rgb}{0.66, 0.66, 0.66}
\makeatletter 
\pretocmd\@bibitem{\color{black}\csname keycolor#1\endcsname}{}{\fail}
\newcommand \citecolor[1]{\@namedef{keycolor#1}{\color{red}}}
\makeatother

\begin{document}
%

\title{\textcolor{black}{Boundary-aware Prototype-driven Adversarial Alignment for Cross-Corpus EEG Emotion Recognition}}
%
%
%
%

\author{\IEEEauthorblockN{
Guangli Li,
Canbiao Wu,
Na Tian,
Li Zhang\textsuperscript{*}, and
Zhen Liang\textsuperscript{*}}\\
\medskip

\IEEEcompsocitemizethanks{
\IEEEcompsocthanksitem Guangli Li, Canbiao Wu, and Na Tian are with the School of Biological Science and Medical Engineering, Hunan University of Technology, Zhuzhou 412008, China. E-mail: guangli010@hut.edu.cn, wucanbiao@m.scnu.edu.cn, and tiann7@mail2.sysu.edu.cn.
\\
\IEEEcompsocthanksitem Li Zhang and Zhen Liang are with the School of Biomedical Engineering, Health Science Center, Shenzhen University, Shenzhen 518060, China, and also with the Guangdong Provincial Key Laboratory of Biomedical Measurements and Ultrasound Imaging, Shenzhen 518060, China, and Zhen Liang also with the Shenzhen Pengrui Brain Science Technology, Shenzhen  518060, China. E-mail: \{zhang, janezliang\}@szu.edu.cn.
\\
\IEEEcompsocthanksitem  \textsuperscript{*}Corresponding author: Li Zhang and Zhen Liang.}}
\IEEEtitleabstractindextext{
\begin{abstract}
\textcolor{black}{Electroencephalography (EEG)-based emotion recognition suffers from severe performance degradation when models are transferred across heterogeneous datasets due to physiological variability, experimental paradigm differences, and device inconsistencies. Existing domain adversarial methods primarily enforce global marginal alignment and often overlook class-conditional mismatch and decision boundary distortion, limiting cross-corpus generalization. In this work, we propose a unified Prototype-driven Adversarial Alignment (PAA) framework for cross-corpus EEG emotion recognition. The framework is progressively instantiated in three configurations: PAA-L, which performs prototype-guided local class-conditional alignment; PAA-C, which further incorporates contrastive semantic regularization to enhance intra-class compactness and inter-class separability; and PAA-M, the full boundary-aware configuration that integrates dual relation-aware classifiers within a three-stage adversarial optimization scheme to explicitly refine controversial samples near decision boundaries. By combining prototype-guided subdomain alignment, contrastive discriminative enhancement, and boundary-aware aggregation within a coherent adversarial architecture, the proposed framework reformulates emotion recognition as a relation-driven representation learning problem, reducing sensitivity to label noise and improving cross-domain stability. Extensive experiments on SEED, SEED-IV, and SEED-V demonstrate state-of-the-art performance under four cross-corpus evaluation protocols, with average improvements of 6.72\%, 5.59\%, 6.69\%, and 4.83\%, respectively. Furthermore, the proposed framework generalizes effectively to clinical depression identification scenarios, validating its robustness in real-world heterogeneous settings. The source code is available at \textit{https://github.com/WuCB-BCI/PAA}}.
\end{abstract}

}

\maketitle

\IEEEdisplaynontitleabstractindextext

%
\IEEEpeerreviewmaketitle

\IEEEraisesectionheading{
\section{Introduction}
\label{sec:introduction}}
\begin{figure*}[th]
\centering
\subfloat{\includegraphics[width=1\textwidth]{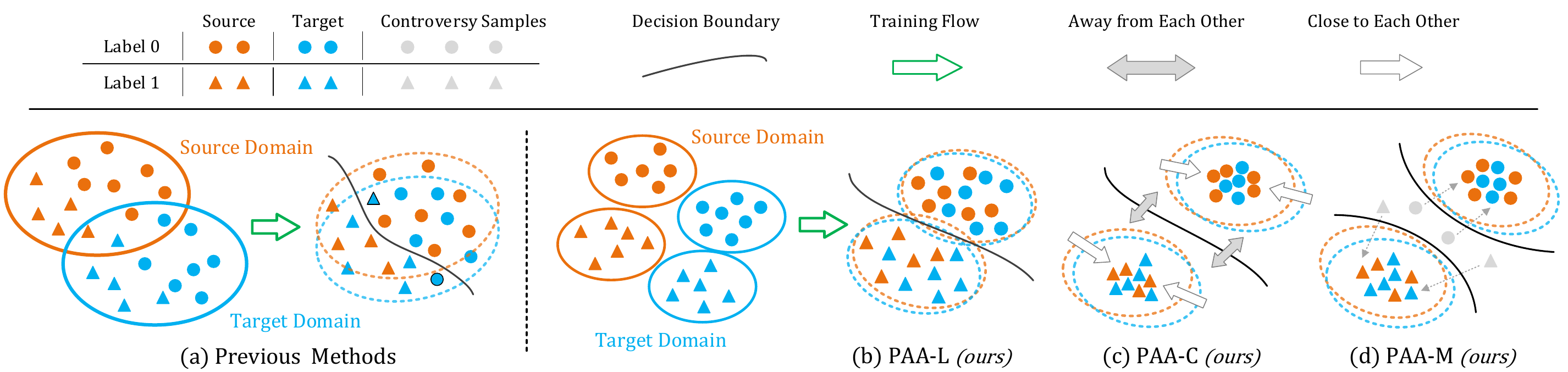}}
\caption{\textcolor{black}{Evolution of cross-domain feature alignment optimization. (a) Traditional global alignment method. (b) Prototype-guided local class-conditional alignment. (c) Contrastive semantic regularization, where intra-class and inter-class optimization are in opposite directions. (d) Maximize boundary-aware aggregation, dual-classifiers are used to identify and aggregate controversy samples.
}}
\label{fig:pr_LMMD_cdd_methods}
\end{figure*}
\IEEEPARstart{E}{} 
\textcolor{black}{lectroencephalography(EEG)-based emotion recognition has become a core topic in affective computing and brain–computer interface systems, owing to its ability to provide direct and objective measurements of neural dynamics \cite{ye2024semi,Ye2023AdaptiveSA}. However, models trained on one EEG dataset often fail to generalize to another due to substantial cross-corpus distribution shifts caused by differences in experimental paradigms, acquisition devices, and subject populations. This cross-corpus discrepancy severely limits the practical deployment of EEG emotion recognition systems. Although domain adaptation techniques attempt to mitigate distribution gaps through adversarial alignment \cite{DANN_2016,Li_Liang_2023}, existing approaches primarily focus on global feature matching and overlook class-conditional inconsistencies, leading to decision boundary distortion and negative transfer in cross-corpus scenarios\cite{ye2024adaptive,liu2025eeg}.} 

\textcolor{black}{Transfer learning has been widely adopted to address inter-subject variability in EEG signals \cite{10509712_2024, Luo2025M3d}. 
Nevertheless, cross-corpus emotion recognition introduces additional challenges beyond cross-subject settings. \textbf{(1) Cross-Corpus Distribution Discrepancy.} Heterogeneous data acquisition protocols and recording environments induce significant feature distribution discrepancies across datasets \cite{liu2025eeg, Liu_2024_eeg}. \textbf{(2) Emotional Label Noise.}  Emotion induction through video stimuli may not perfectly reflect subjects’ internal affective states, inevitably introducing label noise \cite{li2022eeg, liang2021eegfusenet}. \textbf{(3) Decision Boundary Bias under Domain Adversarial Learning.} Conventional domain adversarial learning (DAL) methods enforce global marginal alignment, implicitly assuming that source and target domains share identical class-conditional structures. Such an assumption is often violated in practice and may cause ambiguous decision boundaries, particularly for samples located near inter-class regions.} 

\textcolor{black}{To address these challenges, we propose a unified Prototype-driven Adversarial Alignment (PAA) framework for cross-corpus EEG emotion recognition. The framework is progressively instantiated in three configurations to enable systematic refinement and analysis. The first configuration, termed \textbf{PAA-L}, performs prototype-guided local class-conditional alignment. Instead of enforcing global marginal distribution matching, PAA-L decomposes domain alignment into subdomain-level prototype matching, thereby preserving semantic structure while mitigating coarse distribution discrepancy. Building upon this foundation, the second configuration, \textbf{PAA-C}, augments PAA-L with contrastive semantic regularization. By jointly encouraging intra-class compactness and inter-class separability across domains, PAA-C enhances discriminative consistency and prevents feature collapse during cross-domain adaptation. The full configuration, \textbf{PAA-M}, further extends the framework with a maximize boundary-aware aggregation mechanism embedded within a dual-classifier adversarial architecture. In PAA-M, the feature extractor and domain discriminator form the primary adversarial pair to reduce inter-domain discrepancy, while two structurally identical but optimization-divergent Relation-aware Learning (RaL) classifiers are introduced to amplify classification inconsistency for controversial samples near decision boundaries. Through a three-stage training procedure (representation learning, discrepancy maximization, and boundary refinement), the framework progressively refines cross-domain feature alignment and explicitly alleviates decision boundary bias. By integrating prototype-guided alignment, contrastive regularization, and boundary-aware refinement within a unified adversarial system, the proposed PAA framework reformulates emotion recognition as a relation-driven representation learning problem. This design reduces reliance on potentially noisy emotion labels and improves generalization across heterogeneous EEG corpora.}
Overall, the main contributions of this paper are summarized as follows:
\begin{itemize}
    \item We propose a unified Prototype-driven Adversarial Alignment (PAA) framework for cross-corpus EEG emotion recognition, which reformulates domain adaptation from conventional global marginal matching to class-conditional and boundary-aware feature refinement.
    \item The PAA framework is progressively instantiated in three configurations (PAA-L, PAA-C, and PAA-M) corresponding to local class-conditional alignment, contrastive semantic enhancement, and boundary-aware aggregation, respectively. This progressive design enables systematic refinement of cross-domain consistency from subdomain alignment to discriminative regularization and finally to decision boundary stabilization.
    \item The full configuration, PAA-M, introduces a dual relation-aware classifier architecture with a three-stage adversarial optimization scheme that explicitly identifies and refines controversial samples near inter-class regions, thereby mitigating decision boundary bias and negative transfer.
    \item Extensive evaluations on SEED, SEED-IV, and SEED-V demonstrate state-of-the-art performance under multiple cross-corpus protocols. Moreover, the proposed PAA framework generalizes effectively to clinical depression identification, validating its robustness in real-world heterogeneous scenarios.
\end{itemize}

\section{Related Work}
\label{sec:related_work}
\subsection{Experience-Driven Feature Modeling for EEG Emotion Analysis}
\textcolor{black}{\indent Early studies on EEG emotion recognition primarily relied on handcrafted feature modeling based on domain expertise. Differential Entropy (DE) \cite{Duan_Zhu_2013} was widely adopted to characterize spectral energy distributions associated with affective states. Subsequent works explored information-theoretic feature selection \cite{zhang2025eeg}, manifold-based domain adaptation alignment with dynamic distribution \cite{Luo2025M3d}, and multi-band frequency decomposition using discrete wavelet transform \cite{Baz_2019} to enhance discriminative capability. While experience-driven feature modeling can capture emotion-related signal characteristics within a controlled dataset, these approaches depend heavily on dataset-specific statistical assumptions and manually designed representations. Under cross-subject or cross-corpus settings, distribution shifts caused by heterogeneous recording protocols and subject variability significantly degrade their performance. Consequently, handcrafted modeling lacks robustness for large-scale cross-corpus EEG emotion analysis.}

\subsection{Adversarial Domain Alignment under Cross-Corpus EEG Distribution Shift}
\textcolor{black}{\indent To mitigate inter-subject and inter-dataset discrepancy, adversarial domain alignment has been widely applied to EEG emotion recognition. Representative approaches include multi-adversarial training \cite{JU_2025}, adversarial domain generalization \cite{Gideon_McInnis_Provost_2021}, graph-based adversarial contrastive learning \cite{ye2024semi}, and adversarial temporal convolution networks \cite{HE_2022}. 
Conditional alignment strategies \cite{cheng2025conditional} further attempt to reduce intra-class discrepancy while enlarging inter-class separation. Despite their effectiveness, most adversarial methods focus on aligning global marginal feature distributions across domains. Such alignment implicitly assumes consistent class-conditional structures between source and target datasets. However, in cross-corpus EEG scenarios, heterogeneous acquisition environments and subject populations often induce class-conditional mismatch. When global alignment is enforced under such mismatch, samples near inter-class regions may be incorrectly aligned, resulting in distorted decision boundaries and negative transfer. Therefore, purely marginal adversarial alignment is insufficient for stable cross-corpus generalization.}

\subsection{Prototype-based Semantic Structure Modeling in EEG Emotion Recognition}
\textcolor{black}{\indent Prototype-based modeling encodes each emotion class through representative centroids in embedding space, providing structured semantic supervision. Recent studies have incorporated prototype learning into EEG analysis, including semi-supervised cross-subject modeling \cite{zhou2024eegmatch}, multiscale prototype calibration for neurological disorder identification \cite{Qiu_2024}, and graph-based multi-source prototype representation \cite{GUO_2024}. The PR-PL framework \cite{Li_Liang_2023} further leveraged prototypical pairwise learning to capture intrinsic emotional structure and improve robustness to noisy labels. Although prototype-based methods enhance semantic consistency, most existing approaches integrate prototypes with global distribution alignment. As illustrated in Fig.~\ref{fig:pr_LMMD_cdd_methods} (a), such global prototype alignment may overlook fine-grained subdomain discrepancies and does not explicitly address instability near decision boundaries. Without boundary-aware refinement, cross-corpus adaptation remains vulnerable to decision boundary distortion. These observations suggest that effective cross-corpus EEG emotion recognition requires not only prototype-driven semantic structure modeling but also boundary-aware adversarial alignment to jointly address class-conditional mismatch and boundary instability.}

\begin{table}[ht]
\centering
\caption{Frequently used notations and descriptions.}
\label{tab:T1}
\color{black}
\scalebox{1.1}{
\begin{tabular}{cc}
\toprule
 Notation                       & Description \\
\midrule
$\mathcal{S} / \mathcal{T} $    & Source / Target Domain \\
$D\left(\cdot\right)$           & Domain Discriminator \\ 
$f\left(\cdot\right)$           & Feature Extractor \\
$\phi\left(\cdot\right)$        & Relational Aware \\
$\varphi(\cdot)$                & Feature Mapping into RKHS \\
$\mathcal{P}$                   & Prototype \\
$\mathcal{H}$                   & Reproducing Kernel Hillbert Space (RKHS) \\
$z$                             & Embedding Representation \\
\bottomrule
\end{tabular}
}
\end{table}
\begin{figure*}[ht]
\centering
\subfloat{\includegraphics[width=1\textwidth]{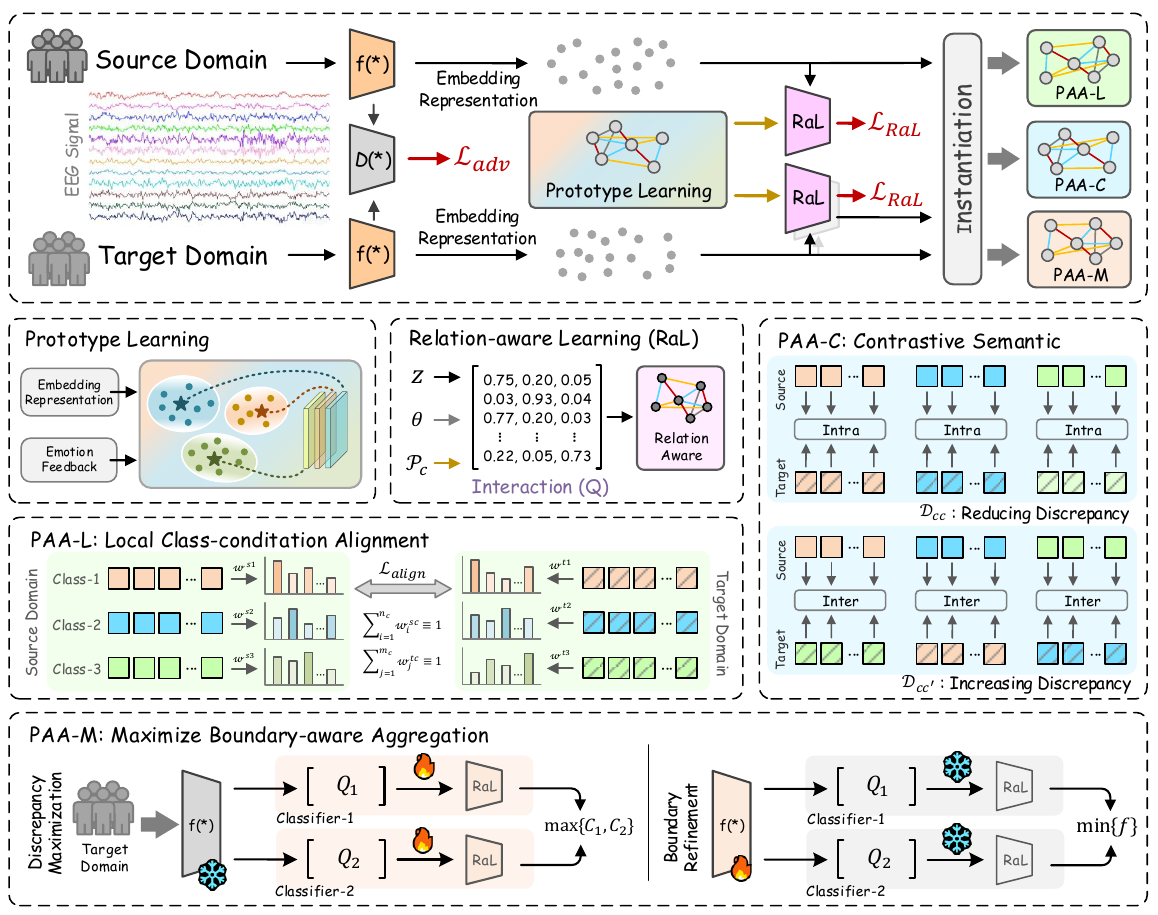}}
\caption{\textcolor{black}{Unified PAA framework. Here, prototype learning is implemented by Eq.~\ref{Eq:proto}; the Relation-aware Learning is implemented by Eq.~\ref{Eq:sim}. The PAA framework instantiates three progressive configurations: PAA-L performs prototype-guided local class-conditional alignment; PAA-C encourages cross-domain intra-class compactness and cross-domain inter-class separability by contrasting semantic regularization; PAA-M is a fully boundary-aware configuration that integrates the maximize boundary-aware aggregation in the three-stage adversarial optimization scheme.}
}
\label{fig:PAA framework}
\end{figure*}

\section{Methodology}
\label{sec:Methodology}

\textcolor{black}{\indent Suppose the labeled source domain $\mathcal{S}$ and unlabeled target domain $\mathcal{T}$ are denoted as:}
\begin{equation}   
    \mathcal{S}=\{(x_i^s,y_i^s)\}_{i=1}^{n},
    \quad
    \mathcal{T}=\{x_j^t\}_{j=1}^{m},
\end{equation} 
where $n$ and $m$ denote the numbers of samples in the source and target domains, respectively. Let $f(\cdot)$ denote the feature extractor and define the embedding representation as:
\begin{equation} 
    z_i^s = f(x_i^s),
    \quad 
    z_j^t = f(x_j^t).
\end{equation} 
Frequently used notations are summarized in TABLE~\ref{tab:T1}.

\subsection{Cross-Corpus Problem Formulation}

\textcolor{black}{\indent Suppose that $P_S$ and $P_T$ represent the distribution of samples in the source and target domains. Under cross-corpus settings, the joint distributions differ:}
\begin{equation}   
    P_S(X,Y) \neq P_T(X,Y),
\end{equation} 
which induces both marginal and class-conditional distribution shifts. The goal of cross-corpus EEG emotion recognition is to learn a representation function $f(\cdot)$ such that the induced conditional distribution becomes invariant:
\begin{equation}  
    P_S(Y \mid f(X)) \approx P_T(Y \mid f(X)).
\end{equation} 
However, empirical risk minimization on $\mathcal{S}$ often results in decision boundary distortion on $\mathcal{T}$. To reduce marginal distribution discrepancy, as shown in Fig.~\ref{fig:PAA framework}, feature extractor $f(\cdot)$ and a domain discriminator $D(\cdot)$ are jointly optimized in an adversarial manner as:
\begin{equation}
\label{Eq:L_adv}
    \mathcal{L}_{adv}=
    -\frac{1}{n}\sum_{i=1}^{n} \log D(z_i^s)
    -\frac{1}{m}\sum_{j=1}^{m} \log \left(1-D(z_j^t)\right).
\end{equation} 
Minimizing $\mathcal{L}_{adv}$ with respect to $D$ and maximizing it with respect to $f$ encourages domain-invariant representations.

However, global alignment does not preserve semantic structure. We therefore introduce prototype representations. For each class $c \in \{1,\dots,C\}$, the prototype is defined as:
\begin{equation}
\label{Eq:proto}
    \mathcal{P}_c=\frac{1}{|\mathcal{S}_c|}
    \sum_{x_i^s \in \mathcal{S}_c} z_i^s,
\end{equation}
where $\mathcal{S}_c$ denotes source samples of class $c$. The prototype can be interpreted as the centroid of class-conditional embeddings.

\indent To further capture relational information between samples, we introduce a Relation-aware Learning (RaL) module. The interaction $Q$ between prototypes and embedding representation is captured by $Q=(z\cdot\mathcal{P}_c\cdot\theta)$, where $\theta$ is a trainable transformation matrix. Here, the relational structure among samples is
\begin{equation}
\label{Eq:sim}
    \phi(z_i,z_j)=
    \frac{Q_i^\top Q_j}{\|Q_i\|_2 \|Q_j\|_2}.
\end{equation}
And the RaL objective is then defined as:
\begin{equation}
\label{Eq:L_RaL}
\begin{aligned}
    \mathcal{L}_{RaL}&=\\ \frac{1}{|N|}\sum_{i,j\in N}&
    \Big[-\mu_{ij}\log \phi(z_i,z_j) -(1-\mu_{ij})\log(1-\phi(z_i,z_j)) \Big]
\end{aligned}
\end{equation}
where $N$ denote the number of samples, $\mu_{ij}$ indicates whether the pair belongs to the same class. This establishes the representation-level foundation for cross-corpus alignment.

\subsection{Prototype-driven Adversarial Alignment}
\label{sec:lmmd}

\textcolor{black}{\indent Although adversarial training reduces marginal distribution discrepancy, it does not guarantee alignment of class-conditional structures across domains. Global alignment may even distort decision boundaries when class priors differ. To preserve semantic geometry, we introduce prototype-guided class-conditional alignment based on conditional mean embedding in Reproducing Kernel Hilbert Space (RKHS). Let $\varphi(\cdot)$ denote the feature mapping into a RKHS $\mathcal{H}$ by a gaussian kernel. The global Maximum Mean Discrepancy (MMD) is defined as:}
\begin{equation}
\label{Eq:MMD}
    \mathrm{MMD}^2(P_S,P_T)=
    \left\|\frac{1}{n}\sum_{i=1}^{n} \varphi(z_i^s)-
    \frac{1}{m}\sum_{j=1}^{m} \varphi(z_j^t)\right\|_{\mathcal{H}}^2.
\end{equation} 
MMD measures the distance between marginal feature distributions in RKHS. However, cross-corpus emotion recognition suffers primarily from conditional shift, e.g. $P_S(z \mid y) \neq P_T(z \mid y)$. To explicitly reduce conditional discrepancy,we perform class-conditional alignment. Specifically, as shown in Fig.~\ref{fig:pr_LMMD_cdd_methods} (b), we need to consider the distribution between different categories in different domains as well as the local importance degree.

\indent Instead of aligning global distributions, we perform class-conditional alignment. Specifically, as shown in Fig.~\ref{fig:pr_LMMD_cdd_methods} (b), we need to consider the distribution between different categories in different domains as well as the local importance degree. Therefore, we propose the local-MMD optimization strategy. For each class $c$, $\omega _i^{sc}$ and $\omega _j^{tc}$ indicate the probability value of the samples in the source and target domain belongs to class $c$, respectively. The local class-conditional alignment is defined as:
\begin{equation}
\label{Eq:L_align}
\begin{aligned}
    \mathcal{L}_{align} & = \\ 
    \frac{1}{C}\sum_{c=1}^{C} &
    \Bigg\|\frac{1}{|\mathcal{S}_c|}
    \sum_{x_i^s \in \mathcal{S}_c} 
    \omega_i^{sc}\varphi(z_i^s) 
    -
    \frac{1}{|\mathcal{T}_c|}
    \sum_{x_j^t \in \mathcal{T}_c} 
    \omega_j^{tc}\varphi(z_j^t)
    \Bigg\|_{\mathcal{H}}^2
\end{aligned}
\end{equation}
Since target labels are unavailable,
we estimate $\omega_j^{tc}$ using pseudo-labels derived from prototype similarity. Here, $\sum_{i=1}^{n_c}\omega _i^{sc} \equiv 1$ and $\sum_{j=1}^{m_c}\omega _j^{tc} \equiv 1$. 

According to domain adaptation theory, the expected target risk is upper bounded by
\begin{equation}
\label{Eq:domainAdpTheory}
    \epsilon_T(h)\le\epsilon_S(h)+
    \frac{1}{2} d_{\mathcal{H}\Delta\mathcal{H}}(P_S,P_T)+\lambda^*,
\end{equation} 
where $\epsilon_T(h)$ and $\epsilon_S(h)$ denote the target and source errors, respectively. $d_{\mathcal{H}\Delta\mathcal{H}}$ measures the hypothesis divergence between source and target distributions, and $\lambda^*$ represents the joint optimal risk of the ideal hypothesis. The divergence term $d_{\mathcal{H}\Delta\mathcal{H}}$ is influenced by both marginal and conditional distribution shifts. While adversarial learning primarily reduces marginal discrepancy $P_S(z) \approx P_T(z)$, it does not explicitly control conditional mismatch. In contrast, minimizing $\mathcal{L}_{align}$ enforces local class-conditional alignment, which approximates reducing
\begin{equation}
    \sum_{c=1}^{C}d_{\mathcal{H}}\big(P_S(z \mid y=c), P_T(z \mid y=c)\big),
\end{equation}
thereby directly mitigating conditional distribution divergence. Under covariate shift assumptions, reducing class-conditional discrepancy decreases hypothesis disagreement across domains, which in turn tightens the divergence term in the adaptation bound. 
If $\| z_S^c - z_T^c \|_{\mathcal{H}} \to 0$, then class centroids coincide in feature space, implying preservation of inter-class margins across domains. This ensures that decision boundaries learned on source classes remain geometrically consistent in target space. Therefore, prototype-driven adversarial alignment reduces conditional generalization gap and forms the first progressive configuration (PAA-L).

\begin{algorithm}[t]
\caption{PAA-M Training Implementation.}
\label{fig:mcd_algorithm}
\color{black}
\textbf{Input:} 
Source domain samples $x^s$ with labels and Unlabeled target domain samples $x^t$. The number of the Epoch-Size ($N$) and Batch-Size ($M$).
\\ \textbf{Implementation:}
\begin{algorithmic}[1]    
    \For{$n = 1$ to $N$}:
        \State \textit{\textbf{Stage 1} \textcolor{gray}{\#Representation Learning}}
        \For{$m = 1$ to $M$}:
            \State Obtain $\mathcal{L}_{adv}$ by Eq.\ref{Eq:L_adv}; \textcolor{gray}{\#Adversarial Loss}
            \State Obtain $\mathcal{L}_{RaL}$ by Eq.\ref{Eq:L_RaL}; \textcolor{gray}{\#Relation-aware Learning}
            \State Obtain $\mathcal{L}_{align}$ by Eq.\ref{Eq:L_align}; \textcolor{gray}{\small\#Class-conditional Alig.}
            \State Obtain $\mathcal{L}_{contrast}$ by Eq.\ref{Eq:L_contrast}; \textcolor{gray}{\small\#Semantic Discrepancy}
            \State $\min{\{f,C_1,C_2\}}$ by Eq.\ref{s1}
        \EndFor
        \State \textit{\textbf{Stage 2} \textcolor{gray}{\#Discrepancy Maximization}}
        \State ** Freeze Feature Generator \textit{\textcolor{gray}{\# $f(\cdot)$}}
        \For{$m = 1$ to $M$}:
            \State $\max\{C_1,C_2\}$ by Eq.\ref{s2}
        \EndFor
        \State ** Unfreeze Feature Generator \textit{\textcolor{gray}{\# $f(\cdot)$}}
        \State \textit{\textbf{Stage 3} \textcolor{gray}{\#Boundary Refinement}}
        \State ** Freeze Dual Classifier \textit{\textcolor{gray}{\# $C_1$ and $C_2$}}
        \For{$m = 1$ to $M$}:
        \State $\min\{f\}$ by Eq.\ref{s3}
        \EndFor
        \State ** Unfreeze Dual Classifier \textit{\textcolor{gray}{\# $C_1$ and $C_2$}}
    \EndFor
\end{algorithmic}
\textbf{Output:} Emotional State.
\end{algorithm}

\subsection{Progressive Class-Conditional and Contrastive Refinement}
\label{sec:cdd}

\textcolor{black}{\indent While class-conditional alignment  enforces semantic consistency across domains, it does not explicitly control the relative geometry between different categories. In cross-corpus settings, even if $P_S(z \mid y=c) \approx P_T(z \mid y=c)$, the distances between different class manifolds may still be small. When transferring a decision boundary learned in the source domain, target samples located near overlapping manifolds may lead to negative transfer and boundary ambiguity. To further refine representation geometry, we introduce a contrastive semantic regularization objective that simultaneously minimizes intra-class semantic discrepancy and maximizes inter-class  semantic separation. Let $\mathcal{D}_{cc}$ and $\mathcal{D}_{cc'}$ denote the RKHS distance between the samples of the same and different class, respectively. The contrastive semantic regularization objective is defined as:}
\begin{equation}
\label{Eq:L_contrast}
    \mathcal{L}_{contrast}=
    \frac{1}{C}\sum_{c=1}^{C} \mathcal{D}_{cc}-
    \frac{1}{C(C-1)}\sum_{c\neq c'} \mathcal{D}_{cc'}.
\end{equation} 
The first term reduces intra-class discrepancy across domains, while the second term enlarges inter-class separation. Thus, as shown in Fig.~\ref{fig:pr_LMMD_cdd_methods} (c),  $\mathcal{L}_{contrast}$ promotes intra-class aggregation and inter-class dispersion simultaneously. From a metric learning perspective, it enlarges the minimum distance between same-class manifolds, expressed as $\min_{cc} \mathrm{dist}(\mathcal{M}_c,\mathcal{M}_{c'})$, and enlarges distance between different-class manifolds: $\max_{cc'} \mathrm{dist}(\mathcal{M}_c,\mathcal{M}_{c'}).$
The model continues to be iteratively updated, samples of the same category gradually finely aligned, while samples of different categories gradually move away from each other, which enhances the model's alignment performance for samples near the decision boundary in cross-corpus. This refinement enhances intra-class compactness and inter-class separability, forming the second progressive configuration (PAA-C).

\begin{figure}[t]
\centering
\subfloat{\includegraphics[width=0.5\textwidth]{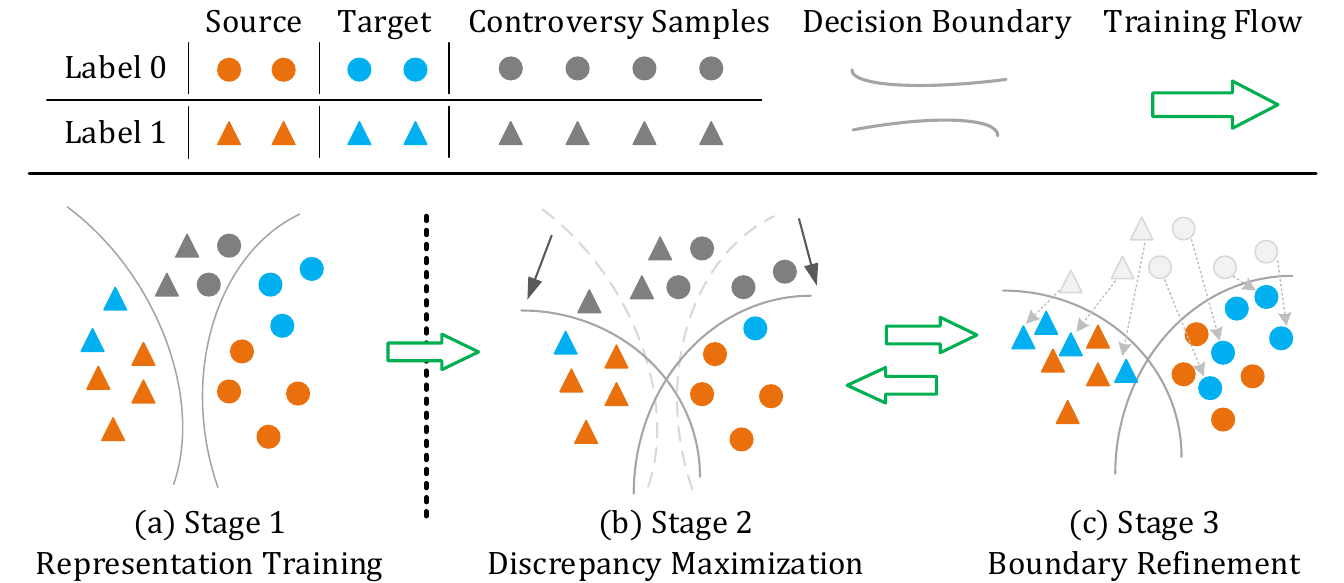}}
\caption{\textcolor{black}{The training process of PAA-M. (a) Firstly, PAA-M learns the base representation. (b) Then, ambiguous regions were identified by maximizing discrepancy, and samples from these regions achieved different results on the two classifiers. (c) Finally, the controversial samples are aggregated.}}
\label{fig:PAA-M_training}
\end{figure}

\subsection{Boundary-Aware Dual-Classifier Optimization}
\label{sec:mcd}

\textcolor{black}{\indent Even with structure-preserving alignment and contrastive refinement, the decision boundary learned from the source domain may remain unstable for target samples near inter-class regions. Such boundary-sensitive samples often exhibit low confidence and high sensitivity to small perturbations. To explicitly model boundary uncertainty, we introduce a dual-classifier adversarial mechanism. Here, we incoporate two classifiers $C_1$ and $C_2$ with identical architectures but independent parameters. Both classifiers have the relation aware ability. Given relational structure of $C_1$ and $C_2$ are $\phi_{c_1}$ and $\phi_{c_2}$, we define classifier discrepancy on target samples as:
\begin{equation}
\label{Eq:L_disc_final}
    \mathcal{L}_{disc}=
    \frac{1}{m^2}\sum_{i,j=1}^{m}\left \| 
    \phi_{c_1}(z_i^t,z_j^t)-\phi_{c_2}(z_i^t,z_j^t) \right \|_1.
\end{equation} 
Large discrepancy indicates samples located in regions where hypotheses disagree, which typically correspond to boundary-sensitive areas. Thus, maximizing $\mathcal{L}_{disc}$ exposes controversial samples near the decision boundary. As given in Eq.~\ref{Eq:domainAdpTheory}, Eq.~(\ref{Eq:L_disc_final}) serves as an empirical approximation of $d_{\mathcal{H}\Delta\mathcal{H}}$. In the other words, maximizing boundary discrepancy identifies target regions where hypotheses diverge, while minimizing controversial samples distribution with respect to $f$ reduces hypothesis disagreement and stabilizes the decision boundary. As shown in Fig.~\ref{fig:pr_LMMD_cdd_methods} (d), This directly reduces the divergence term in the adaptation bound, leading to improved target generalization. This yields the full boundary-aware configuration (PAA-M).
\\ \indent We integrate all components into a unified adversarial optimization objective:
\begin{equation}
\label{Eq:total}
    \min_{f,C}
    \mathcal{L}_{RaL}
    + \lambda_1 \mathcal{L}_{align}
    + \lambda_2 \mathcal{L}_{contrast}
    - \lambda_3 \mathcal{L}_{disc}
    + \mathcal{L}_{adv}.
\end{equation} 
Each progressive configuration corresponds to activating additional regularization terms, systematically tightening the theoretical upper bound of target generalization error. When $\lambda_2=\lambda_3=0$, the framework corresponds to PAA-L. When $\lambda_3=0$ and $\lambda_2>0$, it becomes PAA-C. Activating all terms yields the full boundary-aware configuration (PAA-M). As shown in Fig.~\ref{fig:PAA-M_training}, the optimization of PAA-M alternates between boundary exposure and boundary refinement incorporate three stages: 
\\ \textbf{(1) Representation Learning.} The classifiers and adversarial networks obtain basic emotional representations and extract discriminative features:
\begin{equation}
\label{s1}
    \min_{f,C_1,C_2}
    \mathcal{L}_{RaL}
    + \alpha \mathcal{L}_{align}
    + \beta \mathcal{L}_{contrast}
    + \mathcal{L}_{adv}.
\end{equation}
\textbf{(2) Discrepancy Maximization.} The feature extractor is frozen and classifiers are optimized to maximize disagreement, expands the ambiguous region:
\begin{equation}
\label{s2}
    \max_{C_1,C_2}
    \mathcal{L}_{Ral}-
    \mathcal{L}_{disc}.
\end{equation}
\textbf{(3) Boundary Refinement.} The classifiers are frozen and the feature extractor is optimized to reduce discrepancy, forcing target samples to move away from boundary regions toward confident class clusters:
\begin{equation}
\label{s3}
    \min_{f}
    \mathcal{L}_{RaL}
    +\mathcal{L}_{adv}.
\end{equation}
\indent Overall, these three training steps enable PAA-M to systematically tighten the theoretical upper bound of target risk, which progressively reduces marginal shift, conditional shift, and hypothesis divergence.
}

\begin{table}[]
\begin{center}
\caption{Unified Emotion Label Mapping.}
\label{tab:Label Mapping}
\scalebox{1}{
\begin{tabular}{c|ccc|>{\columncolor{gray!15}}c|>{\columncolor{gray!15}}c}
\toprule
\textit{Number}  & \textit{SEED}    & \textit{SEED-IV}  
                 & \textit{SEED-V}  & \textit{Unified} & \textit{Adopt}\\
\midrule
1  & Positive   & Happy    & Happy     & Positive & \textcolor{orange}{\checkmark} \\
2  & Neutral    & Neutral  & Neutral   & Neutral  & \textcolor{orange}{\checkmark} \\
3  & Negative   & Sad      & Sad       & Negative & \textcolor{orange}{\checkmark} \\
4  & \textcolor{gray}{---}            & \textcolor{lightgray}{Fear}    
   & \textcolor{lightgray}{Fear}       & \textcolor{gray}{---} 
   & \textcolor{gray}{\ding{55}} \\
5  & \textcolor{gray}{---}            & \textcolor{gray}{---}         
   & \textcolor{lightgray}{Disgust}    & \textcolor{gray}{---} 
   & \textcolor{gray}{\ding{55}} \\
\bottomrule
\end{tabular}
}
\end{center}
\end{table}
\begin{table*}[ht]
\begin{center}
\caption{\textcolor{black}{The results in Cross-corpus Cross-subjects Single-session Valuation, expressed as (Acc\% ± Std\%). The dataset combination is represented as (Source Domain $\to$ Target Domain). Here, the results of the baseline model reproduced by us are indicated by '*'.}}
\label{tab:Protocol-1}
\scalebox{0.98}{
\color{black}
\begin{tabular}{lcccccc|>{\columncolor{gray!15}}c}
\toprule
\textit{Methods} 
        & \textit{\scriptsize SEED$\to$SEED-IV}        
        & \textit{\scriptsize SEED$\to$SEED-V}
        & \textit{\scriptsize SEED-IV$\to$SEED}        
        & \textit{\scriptsize SEED-IV$\to$SEED-V}
        & \textit{\scriptsize SEED-V$\to$SEED}         
        & \textit{\scriptsize SEED-V$\to$SEED-IV}
        & \textit{Mean-Acc.}\\ 
\midrule
SVM*\cite{SVM}     
        & 43.14 ± 8.36  & 39.08 ± 9.08  & 44.77 ± 8.62 
        & 31.44 ± 8.56  & 37.27 ± 7.00  & 35.97 ± 7.27  & 38.75 ± 8.15 \\
K-Means*\cite{K_Means} 
        & 48.57 ± 8.62  & 45.33 ± 8.62  & 50.87 ± 6.35
        & 43.37 ± 7.74  & 42.70 ± 6.65  & 46.21 ± 7.60  & 46.18 ± 7.60 \\
KNN*\cite{KNN}    
        & 41.35 ± 4.63  & 36.34 ± 4.52  & 35.09 ± 4.30
        & 26.02 ± 4.44  & 30.63 ± 4.34  & 35.63 ± 4.45  & 34.18 ± 4.45 \\
RF*\cite{breiman2001random} 
        & 43.26 ± 8.83  & 44.06 ± 9.19  & 48.36 ± 8.59
        & 39.99 ± 9.00  & 44.07 ± 7.94  & 31.90 ± 8.22  & 41.94 ± 8.63 \\
KPCA*\cite{KPCA1999}
        & 46.20 ± 8.83  & 47.08 ± 9.31  & 53.21 ± 8.62
        & 34.40 ± 8.67  & 37.12 ± 7.63  & 31.88 ± 7.96  & 41.65 ± 8.50 \\
CORAL*\cite{CORAL2016}  
        & 48.58 ± 8.37  & 44.24 ± 8.62  & 45.30 ± 7.50    
        & 33.90 ± 8.04  & 36.22 ± 8.44  & 35.11 ± 8.00  & 40.56 ± 8.16 \\
SA*\cite{SA2013}  
        & 44.56 ± 8.30  & 42.25 ± 8.39  & 43.12 ± 7.81
        & 35.29 ± 8.18  & 34.78 ± 8.46  & 36.07 ± 8.01  & 39.35 ± 8.19 \\
TCA*\cite{TCA2010}  
        & 43.12 ± 9.14  & 43.25 ± 9.24  & 45.00 ± 8.88
        & 33.46 ± 7.99  & 43.69 ± 7.56  & 38.51 ± 7.81  & 41.17 ± 8.44 \\
GFK*\cite{GFK2012}  
        & 45.76 ± 8.65  & 37.96 ± 8.74  & 44.74 ± 9.00               
        & 33.46 ± 8.21  & 35.40 ± 7.64  & 37.35 ± 7.25  & 39.11 ± 8.25 \\
LeNet*\cite{LeNet}    
        & 42.30 ± 8.55  & 35.60 ± 8.63  & 51.61 ± 8.11               
        & 33.51 ± 7.08  & 41.26 ± 7.15  & 35.92 ± 8.03  & 40.03 ± 7.93 \\
DANN*\cite{DANN_2016}    
        & 44.13 ± 7.85  & 44.92 ± 8.85  & 51.07 ± 8.25 
        & 27.78 ± 8.52  & 41.53 ± 7.13  & 29.93 ± 7.17  & 39.89 ± 7.96 \\
DCORAL*\cite{DCORAL_2016}    
        & 41.48 ± 8.50  & 46.65 ± 9.20  & 54.12 ± 8.36              
        & 33.99 ± 8.20  & 44.10 ± 6.80  & 32.30 ± 8.38  & 42.11 ± 8.24 \\
DAN*\cite{He2018DAN}
        & 44.53 ± 8.56  & 45.91 ± 8.62  & 48.92 ± 8.66
        & 42.95 ± 7.21  & 44.46 ± 7.26  & 46.61 ± 5.20  & 45.56 ± 7.59 \\
DDC*\cite{2014Deep}
        & 43.34 ± 6.18  & 43.66 ± 8.08  & 52.61 ± 8.53
        & 42.82 ± 4.09  & 39.40 ± 6.06  & 42.74 ± 4.94  & 44.10 ± 6.31 \\
DDA*\cite{DDA_2022}    
        & 48.28 ± 8.31  & 46.01 ± 9.33  & 56.26 ± 4.96
        & 47.29 ± 9.11  & 44.26 ± 4.97  & 41.22 ± 7.70  & 47.22 ± 7.40 \\
DMMR*\cite{wang2024dmmr}
        & 49.26 ± 8.14  & 52.23 ± 8.32  & 53.70 ± 8.50
        & 38.15 ± 8.17  & 45.93 ± 7.17  & 39.93 ± 7.03  & 46.53 ± 7.89 \\
BLFBA*\cite{BLFBA_2024}    
        & 46.24 ± 9.03  & 40.95 ± 8.90  & 52.17 ± 6.82  
        & 34.79 ± 8.73  & 44.52 ± 8.12  & 37.12 ± 8.04  & 42.63 ± 8.27 \\
PR-PL*\cite{Li_Liang_2023}    
        & \underline{51.08} ± 5.54	    & \underline{52.71} ± 9.29	
        & \underline{58.53} ± 9.96      & \underline{47.57} ± 6.31	
        & \underline{55.53} ± 11.6	    & \underline{49.44} ± 7.70  
        & \underline{52.48} ± 8.41 \\
SimCLR \cite{chen2020simple}
        & 46.89 ± 13.4  & ---           & 47.27 ± 8.44  
        & ---           & ---           & ---           & 47.08 ± 10.9 \\
TS-TCC \cite{Eldele2021Time}
        & 49.43 ± 9.44  & ---           & 55.38 ± 11.6  
        & ---           & ---           & ---           & 52.41 ± 10.5 \\
\midrule
\multirow{2}{*}{{\normalsize PAA-L} (\textit{ours})} 
        & 51.42 ± 6.98	& \textbf{58.77 ± 5.35}	 & 59.15 ± 6.94	
        & 49.93 ± 4.24	& 59.08 ± 7.83	& 51.37 ± 7.93  & 54.95 ± 6.55 \\
        & \textcolor{orange}{\scriptsize(+0.34)}      
        & \textcolor{orange}{\scriptsize(+6.06)}
        & \textcolor{orange}{\scriptsize(+0.62)}      
        & \textcolor{orange}{\scriptsize(+2.36)}
        & \textcolor{orange}{\scriptsize(+3.55)}      
        & \textcolor{orange}{\scriptsize(+1.93)}
        & \textcolor{orange}{\scriptsize(+2.47)} \\
\cmidrule(lr){2-8}
\multirow{2}{*}{{\normalsize PAA-C} (\textit{ours})}
        & 53.24 ± 6.79	& 53.92 ± 8.31	& 64.91 ± 8.35
        & 51.74 ± 6.78	& 63.74 ± 7.51	& 51.93 ± 7.02  & 56.58 ± 7.46 \\
        & \textcolor{orange}{\scriptsize(+2.16)}       
        & \textcolor{orange}{\scriptsize(+1.21)} 
        & \textcolor{orange}{\scriptsize(+6.38)}       
        & \textcolor{orange}{\scriptsize(+4.17)}
        & \textcolor{orange}{\scriptsize(+8.21)}       
        & \textcolor{orange}{\scriptsize(+2.49)}  
        & \textcolor{orange}{\scriptsize(+4.10)} \\
\cmidrule(lr){2-8}
\multirow{2}{*}{{\normalsize PAA-M} (\textit{ours})}
        & \textbf{53.50 ± 5.74}	                 & 55.97 ± 6.10	
        & \textbf{67.92 ± 7.64}                  & \textbf{54.69 ± 4.09}	
        & \textbf{68.75 ± 6.03}	                 & \textbf{54.38 ± 6.85} 
        & \textbf{59.20 ± 6.08} \\
        & \textcolor{orange}{\scriptsize(+2.42)}      
        & \textcolor{orange}{\scriptsize(+3.26)}
        & \textcolor{orange}{\scriptsize(+9.39)}      
        & \textcolor{orange}{\scriptsize(+7.12)}
        & \textcolor{orange}{\scriptsize(+13.2)}      
        & \textcolor{orange}{\scriptsize(+4.94)}
        & \textcolor{orange}{\scriptsize(+6.72)} \\
\bottomrule
\end{tabular}
}
\end{center}
\end{table*}
\begin{table*}[t]
\begin{center}
\caption{\textcolor{black}{The results in Cross-corpus Cross-subjects Cross-session Valuation, expressed as (Acc\% ± Std\%). The dataset combination is represented as (Source Domain $\to$ Target Domain). Here, the results of the baseline model reproduced by us are indicated by '*'.}}
\label{tab:Protocol-2}
\scalebox{0.98}{
\color{black}
\begin{tabular}{lcccccc|>{\columncolor{gray!15}}c}
\toprule
\textit{Methods} 
        & \textit{\scriptsize SEED$\to$SEED-IV}        
        & \textit{\scriptsize SEED$\to$SEED-V}
        & \textit{\scriptsize SEED-IV$\to$SEED}        
        & \textit{\scriptsize SEED-IV$\to$SEED-V}
        & \textit{\scriptsize SEED-V$\to$SEED}         
        & \textit{\scriptsize SEED-V$\to$SEED-IV}
        & \textit{Mean-Acc.}\\ 
\midrule
SVM*\cite{SVM}     
        & 42.96 ± 8.82  & 40.07 ± 8.07  & 41.23 ± 9.18
        & 38.21 ± 9.24  & 42.82 ± 8.17  & 38.08 ± 7.39  & 40.56 ± 8.48 \\
K-Means*\cite{K_Means} 
        & 47.92 ± 8.02  & 44.21 ± 8.87  & 51.01 ± 6.91
        & 44.42 ± 8.06  & 45.77 ± 6.87  & 46.95 ± 8.00  & 46.71 ± 7.79 \\
KNN*\cite{KNN}    
        & 41.94 ± 4.65  & 43.24 ± 4.41  & 40.92 ± 4.28
        & 38.99 ± 4.39  & 34.92 ± 4.33  & 36.15 ± 4.43  & 39.36 ± 4.42 \\
RF*\cite{breiman2001random} 
        & 47.04 ± 8.40	& 46.09 ± 8.52	& 55.64 ± 7.13	
        & 44.19 ± 8.60	& 41.94 ± 8.28	& 37.60 ± 8.46	& 45.42 ± 8.23 \\
KPCA*\cite{KPCA1999}
        & 51.57 ± 8.81	& 50.73 ± 8.68	& 56.79 ± 8.01	
        & 39.51 ± 8.80	& 44.51 ± 7.51	& 37.17 ± 6.56	& 46.88 ± 8.06 \\
CORAL*\cite{CORAL2016}  
        & 50.08 ± 8.82	& 43.53 ± 8.28	& 42.82 ± 8.31	
        & 36.77 ± 7.98	& 42.83 ± 8.09	& 38.52 ± 7.60	& 42.43 ± 8.18 \\
SA*\cite{SA2013}  
        & 44.80 ± 7.94	& 41.89 ± 8.19	& 49.00 ± 8.12	
        & 34.57 ± 7.90	& 36.10 ± 8.02	& 30.39 ± 8.48	& 39.46 ± 8.11 \\
TCA*\cite{TCA2010}  
        & 47.46 ± 8.71	& 48.90 ± 8.79	& 48.37 ± 7.75	
        & 33.98 ± 8.69	& 51.59 ± 8.52	& 36.57 ± 8.03	& 44.48 ± 8.42 \\
GFK*\cite{GFK2012}  
        & 48.12 ± 8.72	& 45.73 ± 8.51	& 55.69 ± 7.67	
        & 40.24 ± 8.22	& 43.36 ± 8.37	& 36.86 ± 7.23	& 45.00 ± 8.12 \\
LeNet*\cite{LeNet}    
        & 46.49 ± 8.58  & 39.75 ± 6.55  & 47.21 ± 8.68
        & 38.87 ± 7.82  & 46.85 ± 8.44  & 41.27 ± 8.03  & 43.40 ± 8.02 \\
DANN*\cite{DANN_2016}  
        & 45.60 ± 9.30  & 48.04 ± 8.21  & 51.18 ± 8.05
        & 39.50 ± 8.24  & 49.60 ± 9.08  & 46.89 ± 7.68  & 46.80 ± 8.42 \\
DCORAL*\cite{DCORAL_2016}
        & 42.67 ± 8.41  & \underline{54.84} ± 8.46      & 41.85 ± 8.07
        & 46.42 ± 7.91  & \underline{62.16} ± 7.18      & 47.14 ± 7.72  
        & 49.18 ± 7.96 \\
DAN*\cite{He2018DAN}
        & 54.62 ± 8.65	& 52.33 ± 7.49	& 49.69 ± 8.31	
        & 44.56 ± 8.41	& 45.03 ± 9.50	& 43.44 ± 7.79	& 48.45 ± 8.36 \\
DDC*\cite{2014Deep}
        & 42.63 ± 7.59	& 44.14 ± 7.59	& 44.69 ± 8.46	
        & 41.71 ± 7.01	& 37.99 ± 4.39	& 39.35 ± 6.09	& 41.75 ± 6.86 \\
DDA*\cite{DDA_2022}    
        & 45.38 ± 8.51  & 45.87 ± 8.97  & 54.39 ± 4.98
        & 42.65 ± 7.26  & 48.46 ± 8.62  & 44.06 ± 8.24  & 48.45 ± 7.86 \\
DMMR*\cite{wang2024dmmr}
        & \underline{54.97} ± 8.28	    & 48.61 ± 8.43	& 55.26 ± 8.22	
        & 36.12 ± 7.93	& 48.94 ± 8.46	& 40.35 ± 7.62	& 47.38 ± 8.16 \\
BLFBA*\cite{BLFBA_2024}
        & 49.58 ± 8.59  & 41.42 ± 8.58  & 52.12 ± 8.36
        & 45.56 ± 8.11  & 46.34 ± 8.81  & 45.26 ± 7.63  & 46.71 ± 8.35 \\
PR-PL*\cite{Li_Liang_2023}    
        & 52.86 ± 8.42	                & 54.05 ± 6.32	
        & \underline{56.93} ± 7.33      & \underline{46.97} ± 8.60
        & 59.39 ± 8.47	                & \underline{49.66} ± 8.13  
        & \underline{53.31} ± 7.88 \\
\midrule
\multirow{2}{*}{{\normalsize PAA-L} (\textit{ours})}  
        & 54.04 ± 5.41	& 57.06 ± 7.02	& 58.71 ± 7.03
        & 48.76 ± 7.96	& 61.50 ± 7.61	& 50.84 ± 7.76  & 55.15 ± 6.97 \\
        & \textcolor{lightgray}{\scriptsize(-0.93)}     
        & \textcolor{orange}{\scriptsize(+2.22)} 
        & \textcolor{orange}{\scriptsize(+1.78)}        
        & \textcolor{orange}{\scriptsize(+1.79)}
        & \textcolor{lightgray}{\scriptsize(-0.66)}     
        & \textcolor{orange}{\scriptsize(+1.18)} 
        & \textcolor{orange}{\scriptsize(+1.84)} \\
\cmidrule(lr){2-8}
\multirow{2}{*}{{\normalsize PAA-C} (\textit{ours})}
        & 56.35 ± 7.44	& 54.46 ± 7.23	& 58.53 ± 9.93	
        & 47.93 ± 7.63	& 63.18 ± 8.03	& 49.76 ± 7.88  & 55.03 ± 7.86 \\
        & \textcolor{orange}{\scriptsize(+1.38)}        
        & \textcolor{lightgray}{\scriptsize(-0.38)}
        & \textcolor{orange}{\scriptsize(+1.60)}        
        & \textcolor{orange}{\scriptsize(+0.96)}
        & \textcolor{orange}{\scriptsize(+1.02)}        
        & \textcolor{orange}{\scriptsize(+0.10)} 
        & \textcolor{orange}{\scriptsize(+1.72)} \\
\cmidrule(lr){2-8}
\multirow{2}{*}{{\normalsize PAA-M} (\textit{ours})}
        & \textbf{57.40 ± 5.19}	                   & \textbf{58.90 ± 6.57}	
        & \textbf{60.77 ± 5.67}	                   & \textbf{54.07 ± 5.12}	
        & \textbf{68.83 ± 5.74}	                   & \textbf{53.42 ± 7.04}  
        & \textbf{58.90 ± 5.89} \\
        & \textcolor{orange}{\scriptsize(+2.43)}        
        & \textcolor{orange}{\scriptsize(+4.06)} 
        & \textcolor{orange}{\scriptsize(+3.84)}        
        & \textcolor{orange}{\scriptsize(+7.10)} 
        & \textcolor{orange}{\scriptsize(+6.67)}        
        & \textcolor{orange}{\scriptsize(+3.76)} 
        & \textcolor{orange}{\scriptsize(+5.59)} \\
\bottomrule
\end{tabular}
}
\end{center}
\end{table*}

\begin{table*}[t]
\begin{center}
\caption{\textcolor{black}{The results in Cross-corpus Cross-subjects Single-session Leave-one-subject-out Cross-valuation, expressed as (Acc\% ± Std\%). The dataset combination is represented as (Source Domain $\to$ Target Domain). Here, the results of the baseline model reproduced by us are indicated by '*'.}}
\label{tab:Protocol-3}
\scalebox{0.98}{
\color{black}
\begin{tabular}{lcccccc|>{\columncolor{gray!15}}c}
\toprule
\textit{Methods} 
        & \textit{\scriptsize SEED$\to$SEED-IV}        
        & \textit{\scriptsize SEED$\to$SEED-V}
        & \textit{\scriptsize SEED-IV$\to$SEED}        
        & \textit{\scriptsize SEED-IV$\to$SEED-V}
        & \textit{\scriptsize SEED-V$\to$SEED}         
        & \textit{\scriptsize SEED-V$\to$SEED-IV}
        & \textit{Mean-Acc.}\\ 
\midrule
SA*\cite{SA2013} 
        & 39.29 ± 7.20	& 31.01 ± 7.13	& 40.85 ± 6.94
        & 34.46 ± 7.14	& 31.22 ± 7.69	& 33.52 ± 7.26	& 35.06 ± 7.23 \\
TCA*\cite{TCA2010} 
        & 42.36 ± 8.00	& 42.90 ± 8.24	& 47.48 ± 7.91	
        & 34.84 ± 7.82	& 46.94 ± 5.86	& 40.82 ± 7.25	& 42.56 ± 7.51 \\
GFK*\cite{GFK2012}
        & 43.14 ± 8.30	& 38.00 ± 8.05	& 49.82 ± 8.38	
        & 32.54 ± 8.08	& 33.03 ± 6.85	& 34.21 ± 6.48	& 38.46 ± 7.69 \\
LeNet*\cite{LeNet}    
        & 42.03 ± 8.64  & 35.22 ± 8.56  & 51.91 ± 8.20
        & 32.42 ± 7.13  & 42.86 ± 7.34  & 35.24 ± 7.83  & 39.95 ± 7.95 \\
DANN*\cite{DANN_2016}  
        & 40.19 ± 8.09  & 40.37 ± 7.69  & 42.85 ± 7.51
        & 26.73 ± 7.88  & 37.26 ± 6.58  & 37.77 ± 8.06  & 37.53 ± 7.64 \\
DCORAL*\cite{DCORAL_2016}     
        & 41.67 ± 7.94  & 42.30 ± 8.52  & 46.95 ± 7.78
        & 35.59 ± 7.76  & 37.73 ± 7.71  & 36.12 ± 6.93  & 40.06 ± 7.77 \\
DAN*\cite{He2018DAN} 
        & 44.33 ± 8.14	& 33.76 ± 7.88	& 44.59 ± 7.96
        & 43.75 ± 7.54	& 39.06 ± 6.67	& 36.43 ± 7.11	& 40.32 ± 7.55 \\
DDC*\cite{2014Deep} 
        & 39.67 ± 6.77	& 35.62 ± 4.15	& 40.10 ± 7.56	
        & 37.35 ± 7.96	& 36.53 ± 4.90	& 37.21 ± 5.78	& 37.75 ± 6.19 \\
DDA*\cite{DDA_2022}    
        & 35.85 ± 8.47  & 35.03 ± 8.09  & \underline{57.05} ± 8.08   
        & 32.32 ± 8.95  & 43.63 ± 7.91  & 34.03 ± 7.89  & 39.65 ± 8.23 \\
BLFBA*\cite{BLFBA_2024}    
        & 44.23 ± 6.89  & 49.80 ± 8.10  & 51.96 ± 7.58
        & 40.17 ± 4.49  & 47.36 ± 7.29  & \underline{47.19} ± 5.44  
        & 46.79 ± 6.63 \\
PR-PL*\cite{Li_Liang_2023}    
        & \underline{52.62} ± 7.24      & \underline{49.82} ± 10.5  
        & 55.75 ± 14.3                  & \underline{55.78} ± 11.9      
        & \underline{51.07} ± 4.16      & 45.19 ± 8.22  
        & \underline{50.21} ± 9.38 \\
\midrule
\multirow{2}{*}{{\normalsize PAA-L} (\textit{ours})}
        & 48.51 ± 7.85  & 50.66 ± 9.86  & \textbf{58.37 ± 9.07}
        & 56.91 ± 10.2  & 52.39 ± 9.06  & 47.07 ± 12.1  & 52.32 ± 9.86 \\ 
        & \textcolor{lightgray}{\scriptsize(-4.11)}     
        & \textcolor{orange}{\scriptsize(+0.84)} 
        & \textcolor{orange}{\scriptsize(+1.32)}        
        & \textcolor{orange}{\scriptsize(+1.13)} 
        & \textcolor{orange}{\scriptsize(+1.32)}        
        & \textcolor{lightgray}{\scriptsize(-0.12)} 
        & \textcolor{orange}{\scriptsize(+2.11)} \\
\cmidrule(lr){2-8}
\multirow{2}{*}{{\normalsize PAA-C} (\textit{ours})} 
        & 54.09 ± 6.20  & 48.45 ± 13.3  & 56.81 ± 12.8
        & 60.07 ± 13.6  & 51.14 ± 5.77  & 50.34 ± 8.84  & 53.48 ± 10.1 \\
        & \textcolor{orange}{\scriptsize(+1.47)}        
        & \textcolor{lightgray}{\scriptsize(-1.37)} 
        & \textcolor{lightgray}{\scriptsize(-0.24)}     
        & \textcolor{orange}{\scriptsize(+4.29)} 
        & \textcolor{orange}{\scriptsize(+0.07)}        
        & \textcolor{orange}{\scriptsize(+3.15)} 
        & \textcolor{orange}{\scriptsize(+3.27)} \\
\cmidrule(lr){2-8}
\multirow{2}{*}{{\normalsize PAA-M} (\textit{ours})} 
        & \textbf{56.29 ± 6.44}                    & \textbf{57.87 ± 8.20}  
        & 58.26 ± 8.10                             & \textbf{60.87 ± 4.99}  
        & \textbf{55.32 ± 6.20}                    & \textbf{53.79 ± 3.98}  
        & \textbf{56.90 ± 6.32} \\
        & \textcolor{orange}{\scriptsize(+3.67)}        
        & \textcolor{orange}{\scriptsize(+8.05)}
        & \textcolor{orange}{\scriptsize(+1.21)}        
        & \textcolor{orange}{\scriptsize(+5.09)}
        & \textcolor{orange}{\scriptsize(+4.25)}        
        & \textcolor{orange}{\scriptsize(+6.60)}
        & \textcolor{orange}{\scriptsize(+6.69)} \\
\bottomrule
\end{tabular}
}
\end{center}
\end{table*}

\begin{table*}[t]
\begin{center}
\caption{\textcolor{black}{The results in Cross-corpus Cross-subjects Cross-session Leave-one-subject-out Cross-valuation, expressed as (Acc\% ± Std\%). The dataset combination is represented as (Source Domain $\to$ Target Domain). Here, the results of the baseline model reproduced by us are indicated by '*'.}}
\label{tab:Protocol-4}
\scalebox{0.98}{
\color{black}
\begin{tabular}{lcccccc|>{\columncolor{gray!15}}c}
\toprule
\textit{Methods} 
        & \textit{\scriptsize SEED$\to$SEED-IV}        
        & \textit{\scriptsize SEED$\to$SEED-V}
        & \textit{\scriptsize SEED-IV$\to$SEED}        
        & \textit{\scriptsize SEED-IV$\to$SEED-V}
        & \textit{\scriptsize SEED-V$\to$SEED}         
        & \textit{\scriptsize SEED-V$\to$SEED-IV}
        & \textit{Mean-Acc.}\\ 
\midrule
SA*\cite{SA2013}
        & 45.60 ± 7.52	& 39.41 ± 7.54	& 48.21 ± 7.88	
        & 35.50 ± 7.52	& 34.86 ± 7.67	& 32.76 ± 8.19	& 39.39 ± 7.72 \\
TCA*\cite{TCA2010}
        & 48.03 ± 8.59	& 42.23 ± 8.24	& 52.86 ± 8.32
        & 30.12 ± 8.13	& 46.64 ± 8.24	& 35.05 ± 0.76	& 42.49 ± 7.05 \\
GFK*\cite{GFK2012}
        & 49.06 ± 10.3	& 45.86 ± 8.23	& 52.14 ± 7.55	
        & 40.48 ± 7.64	& 42.19 ± 7.96	& 37.45 ± 7.19	& 44.53 ± 7.84 \\
LeNet*\cite{LeNet}    
        & 46.01 ± 8.56  & 41.31 ± 6.66  & 47.01 ± 8.65
        & 39.10 ± 7.68  & 45.03 ± 8.19  & 40.20 ± 7.79  & 43.11 ± 7.92 \\
DANN*\cite{DANN_2016}  
        & 44.10 ± 8.69  & 40.10 ± 8.49  & 47.32 ± 8.22
        & 41.25 ± 7.75  & 48.72 ± 8.52  & 34.35 ± 8.57  & 42.64 ± 8.37 \\
DCORAL*\cite{DCORAL_2016}     
        & 43.76 ± 8.36  & 44.20 ± 7.62  & 43.24 ± 7.89
        & 44.63 ± 7.83  & 44.82 ± 7.40  & 39.95 ± 7.15  & 43.43 ± 7.71 \\
DAN*\cite{He2018DAN} 
        & 49.64 ± 9.55	& 47.56 ± 7.68	& 52.38 ± 7.94	
        & 37.33 ± 7.67	& 40.32 ± 6.91	& 36.69 ± 7.23	& 43.91 ± 7.63 \\
DDC*\cite{2014Deep}
        & 45.05 ± 7.72	& 34.63 ± 5.38	& 39.92 ± 7.25
        & 35.97 ± 6.76	& 34.59 ± 3.80	& 38.17 ± 4.85	& 38.06 ± 6.09 \\
DDA*\cite{DDA_2022}    
        & 44.23 ± 8.48  & 41.11 ± 8.27  & 54.50 ± 8.31
        & 42.09 ± 7.46  & 49.45 ± 9.08  & 43.76 ± 8.14  & 45.86 ± 8.29 \\
BLFBA*\cite{BLFBA_2024}    
        & \underline{50.75} ± 6.00      & 51.29 ± 6.11  
        & 55.81 ± 7.56                  & \underline{52.13} ± 8.75      
        & 49.18 ± 7.30                  & \underline{50.28} ± 6.57  
        & \underline{51.57} ± 7.05 \\
PR-PL*\cite{Li_Liang_2023}    
        & 50.15 ± 5.09                  & \underline{52.16} ± 10.2 
        & \underline{56.39} ± 5.27      & 45.26 ± 7.90  
        & \underline{51.90} ± 3.81      & 48.13 ± 6.73  
        & 50.67 ± 6.50 \\
\midrule
\multirow{2}{*}{{\normalsize PAA-L} (\textit{ours})}  
        & \textbf{54.56 ± 6.83}         & 49.85 ± 10.7  & 56.52 ± 3.97
        & 53.62 ± 7.03  & 57.80 ± 5.75  & 53.05 ± 5.99  & 54.23 ± 6.71 \\
        & \textcolor{orange}{\scriptsize(+3.81)}        
        & \textcolor{lightgray}{\scriptsize(-2.31)}
        & \textcolor{orange}{\scriptsize(+0.13)}        
        & \textcolor{orange}{\scriptsize(+1.49)}
        & \textcolor{orange}{\scriptsize(+5.90)}        
        & \textcolor{orange}{\scriptsize(+2.77)}
        & \textcolor{orange}{\scriptsize(+2.66)} \\
\cmidrule(lr){2-8}
\multirow{2}{*}{{\normalsize PAA-C} (\textit{ours})} 
        & 52.89 ± 5.78  & 51.41 ± 8.40  & 56.55 ± 5.53
        & 51.83 ± 3.75  & 54.71 ± 9.40  & 51.35 ± 5.83  & 53.12 ± 6.45 \\
        & \textcolor{orange}{\scriptsize(+2.14)}        
        & \textcolor{lightgray}{\scriptsize(-0.75)}
        & \textcolor{orange}{\scriptsize(+0.16)}        
        & \textcolor{lightgray}{\scriptsize(-0.30)}
        & \textcolor{orange}{\scriptsize(+2.81)}        
        & \textcolor{orange}{\scriptsize(+1.07)}
        & \textcolor{orange}{\scriptsize(+1.55)} \\
\cmidrule(lr){2-8}
\multirow{2}{*}{{\normalsize PAA-M} (\textit{ours})}  
        & 52.62 ± 6.79                             & \textbf{54.50 ± 8.87}  
        & \textbf{62.93 ± 5.26}                    & \textbf{55.72 ± 3.30}  
        & \textbf{59.12 ± 6.33}                    & \textbf{53.50 ± 5.86}  
        & \textbf{56.40 ± 6.07} \\
        & \textcolor{orange}{\scriptsize(+1.87)}        
        & \textcolor{orange}{\scriptsize(+2.34)}
        & \textcolor{orange}{\scriptsize(+6.54)}        
        & \textcolor{orange}{\scriptsize(+3.59)}
        & \textcolor{orange}{\scriptsize(+7.22)}        
        & \textcolor{orange}{\scriptsize(+3.22)}
        & \textcolor{orange}{\scriptsize(+4.83)} \\
\bottomrule
\end{tabular}
}
\end{center}
\end{table*}

\section{Experimental Results}
\label{sec:experiment}

\subsection{Dataset and Data Preprocessing}

\textcolor{black}{\indent We validated our proposed three cross-corpus optimization strategies on three publicly available datasets (SEED, SEED-IV, and SEED-V \cite{Wei_2015_eeg, pi_2018_uns, Li_2021_eeg}). For each dataset, a total of 15 subjects (16 subjects in SEED-V) participated in the experiment, and each subject completed three sessions on different dates. Each session in SEED consisted of 15 trials, including visual stimuli of three emotions (negative, neutral, and positive). Each session in SEED-IV consisted of 24 trials, including visual stimuli of three emotions (happy, sad, fear, and neutral). Each session in SEED-IV consisted of 15 trials, including visual stimuli of five emotions (happy, neutral, sad, disgust, and fear).}

\textcolor{black}{For consistency and strictness, we define happiness as a positive emotion and sadness as a negative emotion. As shown in TABLE~\ref{tab:Label Mapping}, we only retained positive, neutral, and negative emotions. All the data were uniformly preprocessed. We downsampled EEG signals to 200 Hz and manually removed interfering signals such as EMG and EOG. Subsequently, we used a band-pass filter (0.3-50 Hz) to filter the EEG signal and split it into 1-second non-overlapping segments. Based on the predefined frequency bands (\textit{Delta} 1-3 Hz, \textit{Theta} 4-7 Hz, \textit{Alpha} 8-13 Hz, \textit{Beta} 14-30 Hz and \textit{Gamma} 31-50 Hz), we extract the corresponding differential entropy features to represent the logarithm energy spectrum in a specific frequency band. Significantly, since the source and target domains come from separate datasets, there is no leakage of any information.}

\subsection{Model Implementation}
    \textcolor{black}{\indent The feature extractor $f(\cdot)$ and discriminator $D(\cdot)$ of PAA are constructed using Multilayer Perceptron (MLP) with ReLU activation functions. In PAA-L and PAA-C, Root Mean Square Propagation (RMSProp) optimizer is used to optimize the model. In PAA-M, we combine the Adaptive Moment Estimation and RMSProp optimizers, which show better performance compared to other classical optimizers. The learning rate of 1e-3, with a training epoch size of 300 and a batch size of 256. All models are executed under the following configuration: CUDA 11.6, PyTorch 1.12.1, and a NVIDIA GeForce RTX 3090.}

\subsection{Experiment Protocols}
\label{experiment protocols}
    \textcolor{black}{
    \indent To comprehensively evaluate the robustness and stability of the models, we randomly select two databases as the source domain and target domain, denoted as (Source Domain$\to$Target Domain). Thus, it produces 6 different cross-corpus training combinations. Additionally, we employed four different validation protocols.
    \textbf{(1) Cross-corpus Cross-subjects Single-session Valuation}. 
    This protocol is widely used in EEG-based emotion recognition. The first session of all subjects from one database as the source domain, while the first session of all subjects from another database as the target domain.
    \textbf{(2) Cross-corpus Cross-subjects Cross-session Valuation}. All sessions from all subjects within one database as source domain and the another database as target domain. This evaluation protocol presents significant challenges in cross-corpus emotion recognition.
    \textbf{(3) Cross-corpus Cross-subjects Single-session Leave-one-subject-out (LOSO) Cross-valuation in Unseen-target}. We verified the robustness and generalization performance of the model on unseen data. In the target domain database, the first session of one subject is used as an independent test set, while the remaining subjects will be used as target domain data. The independent test set does not participate in the training and is only used to verify the performance. Each subject in the target domain database is alternately considered as an independent test set, while the source domain remains unchanged.
    \textbf{(4) Cross-corpus Cross-subjects Cross-session Unseen LOSO Cross-valuation in Unseen-target}. This protocol is similar to the protocol-(3). However, we treat all sessions of one subject as independent test sets and the remaining subject as regular target domain data, which imposes stricter validation requirements.}

\subsection{Cross-corpus Cross-subjects Single-session Valuation}
    \textcolor{black}{\indent As shown in TABLE~\ref{tab:Protocol-1}, the results are expressed as (Acc ± Std) and the best baseline model is underlined. The experimental results show that the PAA-L, -C and -M methods are better than the baseline models under six database training combinations, with the average accuracy of 54.95\%, 56.58\% and 59.20\%, respectively. And Compared with the baseline model, the performance has improved by 2.47\%, 4.10\%, and 6.72\% respectively. PAA-C achieves the best performance with 58.77\% accuracy in SEED$\to$SEED-V, while PAA-M achieves the best performance for the remaining database combinations. Overall, PAA-M achieves the state-of-the-art (SOTA) performance. The results show that the proposed three strategies all improve the feature alignment ability and recognition performance near the decision boundary in cross-corpus tasks.}

\subsection{Cross-corpus Cross-subjects Cross-session Valuation}
    \textcolor{black}{\indent This experimental protocol poses higher challenges to the robustness and generalization performance of the model. The results of this experimental protocol are shown in TABLE~\ref{tab:Protocol-2}, the average accuracy of PAA-L, PAA-C and PAA-M methods are 54.95\%, 56.58\% and 59.20\%, respectively, which are better than the baseline model. Specifically, the PAA-M model achieves SOTA performance in all six cross-corpus training combinations, with an average accuracy improvement of 5.59\% over the baseline model. And the highest accuracy in SEED-V$\to$SEED is 68.53\%. In addition, our proposed PAA-L and PAA-C optimization strategies improve 1.84\% and 1.72\%. These results prove that the proposed optimization strategy has excellent generalization ability and potential.}

\subsection{Cross-corpus Cross-subjects Single-session LOSO Cross-valuation in Unseen-target}
    \textcolor{black}{\indent Since the independent test set excluded from model training, this experimental protocols eliminates potential bias contamination, which is more in line with real-world application scenarios, while posing greater challenges to model generalization. As shown in TABLE~\ref{tab:Protocol-3}, the PAA-L model achieves the best performance in SEED-IV$\to$SEED with an accuracy of 58.37\%. PAA-C is slightly lower than the baseline model in SEED$\to$SEED-V and SEED-IV$\to$SEED, but the average accuracy reaches 53.48\%, which is 3.27\% higher than the baseline model. In addition, PAA-M achieves SOTA performance with an average accuracy of 56.9\%, an improvement of 6.69\% over the baseline model. These results demonstrate the potential of the proposed optimization strategy for real-world applications.}

\subsection{Cross-corpus Cross-subjects Cross-session LOSO Cross-valuation in Unseen-target}
    \textcolor{black}{\indent This experimental protocol is the most stringent of all four protocols. The results are shown in TABLE~\ref{tab:Protocol-4}, the PAA-L strategy achieves the best performance with 54.56\% accuracy in SEED$\to$SEED-IV, which is 3.8\% higher than the baseline model. In SEED$\to$SEED-V, PAA-L and PAA-C are slightly lower than the baseline model. However, the average accuracy of PAA-L and PAA-C is 49.85\% and 51.41\%, which are 2.66\% and 1.55\% higher than the baseline model respectively. Overall, the PAA-M model continues to demonstrate the greatest application potential, achieving an average accuracy of 56.40\% in all corpus combinations, slightly higher than the PPA-L and -C strategies. These results demonstrate the improvement of model performance by proposed optimization strategy, which is beneficial for fine-grained alignment of samples near the decision boundary.}

\begin{figure*}[ht]
\centering
\subfloat{\includegraphics[width=1\textwidth]{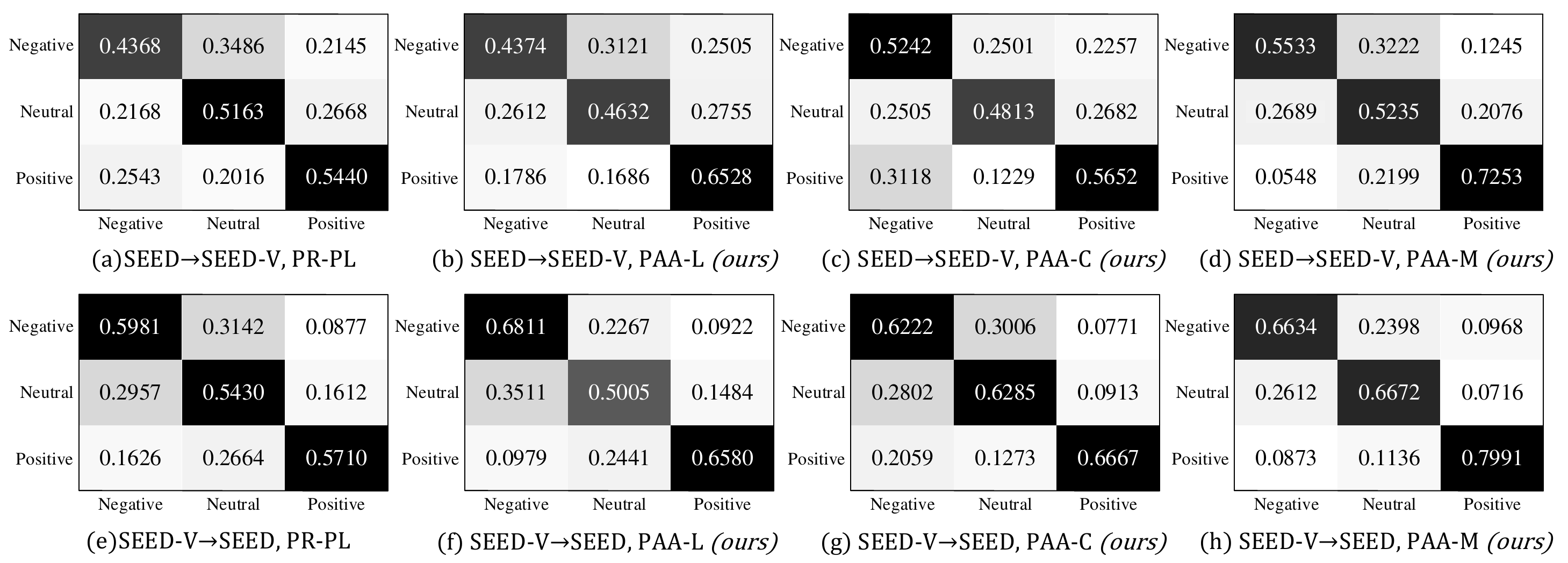}}
\caption{
    \textcolor{black}{Confusion matrices of baseline model PR-PL and proposed PAA-L, PAA-C, and PAA-M. The horizontal axis represents the predicted labels, while the vertical axis represents the true labels.}
    }
\label{fig:confusion_matrix}
\end{figure*}

\begin{table*}[th]
\begin{center}
\caption{\textcolor{black}{The performance comparison of Relation-aware Learning (RaL) and traditional Single-Samples Learning (SSL). Here, NR represents the noise ratio. Gap represents the performance gap between NR of 10\% and 40\%.}
}
\label{tab:noisy_labels_1}
\scalebox{0.98}{
\color{black}
\begin{tabular}{llcccccc|>{\columncolor{gray!15}}c}
\toprule 
        & \textit{NR}
        & \textit{\scriptsize SEED$\to$SEED-IV}        
        & \textit{\scriptsize SEED$\to$SEED-V}
        & \textit{\scriptsize SEED-IV$\to$SEED}        
        & \textit{\scriptsize SEED-IV$\to$SEED-V}
        & \textit{\scriptsize SEED-V$\to$SEED}         
        & \textit{\scriptsize SEED-V$\to$SEED-IV}
        & \textit{Mean-Acc.}\\ 
\midrule
\multirow{4.6}{*}{\textit{\normalsize with}}
&\textit{\textit{10\%}}
        & 55.51 ± 7.20  & 53.02 ± 5.22  & 60.48 ± 9.30
        & 52.01 ± 7.03  & 67.97 ± 7.05  & 53.30 ± 8.10  & 57.04 ± 7.32 \\
\multirow{4.5}{*}{\textit{\normalsize RaL}}
&\textit{\textit{20\%}} 
        & 52.16 ± 6.18  & 52.02 ± 6.15  & 60.40 ± 7.44
        & 50.56 ± 6.09  & 66.16 ± 7.44  & 51.79 ± 6.16  & 55.52 ± 5.57 \\
&\textit{\textit{30\%}} 
        & 51.95 ± 4.65  & 51.81 ± 5.29  & 60.20 ± 8.89
        & 49.34 ± 6.09  & 65.22 ± 3.03  & 50.10 ± 5.38  & 54.77 ± 5.56 \\
&\textit{\textit{40\%}} 
        & 50.03 ± 4.93  & 51.51 ± 7.72  & 59.96 ± 9.61
        & 48.62 ± 5.37  & 64.57 ± 3.11  & 48.76 ± 5.04  & 53.91 ± 5.96 \\
&\textit{\textit{Gap}} 
        & \textcolor{orange}{\scriptsize(-5.48)}     
        & \textcolor{orange}{\scriptsize(-1.51)}
        & \textcolor{orange}{\scriptsize(-0.52)}     
        & \textcolor{orange}{\scriptsize(-3.39)}
        & \textcolor{orange}{\scriptsize(-1.42)}     
        & \textcolor{orange}{\scriptsize(-4.54)}
        & \textcolor{orange}{\scriptsize(-3.13)} \\
\cmidrule(lr){2-9}
\multirow{4.6}{*}{\textit{\normalsize with}}
&\textit{\textit{10\%}} 
        & 48.82 ± 7.74  & 48.90 ± 4.45  & 54.25 ± 6.57
        & 48.39 ± 4.16  & 55.92 ± 6.28  & 45.08 ± 4.13  & 50.23 ± 5.56 \\
\multirow{4.5}{*}{\textit{\normalsize SSL}}
&\textit{\textit{20\%}} 
        & 45.13 ± 5.68  & 47.84 ± 5.85  & 53.01 ± 6.51
        & 45.75 ± 4.99  & 52.99 ± 5.90  & 43.95 ± 4.30  & 48.11 ± 5.54 \\
&\textit{\textit{30\%}} 
        & 43.09 ± 5.95  & 46.03 ± 7.52  & 49.96 ± 7.36
        & 44.34 ± 3.07  & 51.09 ± 6.57  & 42.16 ± 7.18  & 46.11 ± 6.28 \\
&\textit{\textit{40\%}} 
        & 41.67 ± 4.76  & 44.85 ± 4.16  & 48.13 ± 7.03 
        & 43.81 ± 6.10  & 49.51 ± 4.68  & 39.72 ± 3.46  & 44.62 ± 5.03 \\
&\textit{\textit{Gap}} 
        & \textcolor{Cerulean}{\scriptsize(-7.15)}     
        & \textcolor{Cerulean}{\scriptsize(-4.05)}
        & \textcolor{Cerulean}{\scriptsize(-6.12)}     
        & \textcolor{Cerulean}{\scriptsize(-4.58)}
        & \textcolor{Cerulean}{\scriptsize(-6.41)}     
        & \textcolor{Cerulean}{\scriptsize(-5.36)}
        & \textcolor{Cerulean}{\scriptsize(-5.61)} \\
\bottomrule
\end{tabular}
}
\end{center}
\end{table*}
\begin{table}[th]
\begin{center}
\caption{\textcolor{black}{Comparative results of noise tolerance in different strtegies. Here, NR represents the noise ratio.}}
\label{tab:noisy_labels_2}
\scalebox{1}{
\begin{tabularx}{0.95\linewidth}{l *{5}{>{\centering\arraybackslash}X}}
\toprule
\textit{Strategy}    
    & 0\%  & 10\%    & 20\%     & 30\%   & 40\%  \\
\midrule
\textit{with RaL}
    & 58.90     & 57.04     & 55.52     & 54.77     & 53.91 \\
\textit{with SSL}
    & 53.74     & 50.23     & 48.11     & 46.11     & 44.62 \\
\textit{Gap}
    & \textcolor{orange}{\scriptsize(-5.16)}  
    & \textcolor{orange}{\scriptsize(-6.56)} 
    & \textcolor{orange}{\scriptsize(-7.41)}  
    & \textcolor{orange}{\scriptsize(-8.66)} 
    & \textcolor{orange}{\scriptsize(-9.29)} \\
\bottomrule
\end{tabularx}
}
\end{center}
\end{table}

\begin{figure*}[th]
\centering
\subfloat{\includegraphics[width=0.99\textwidth]{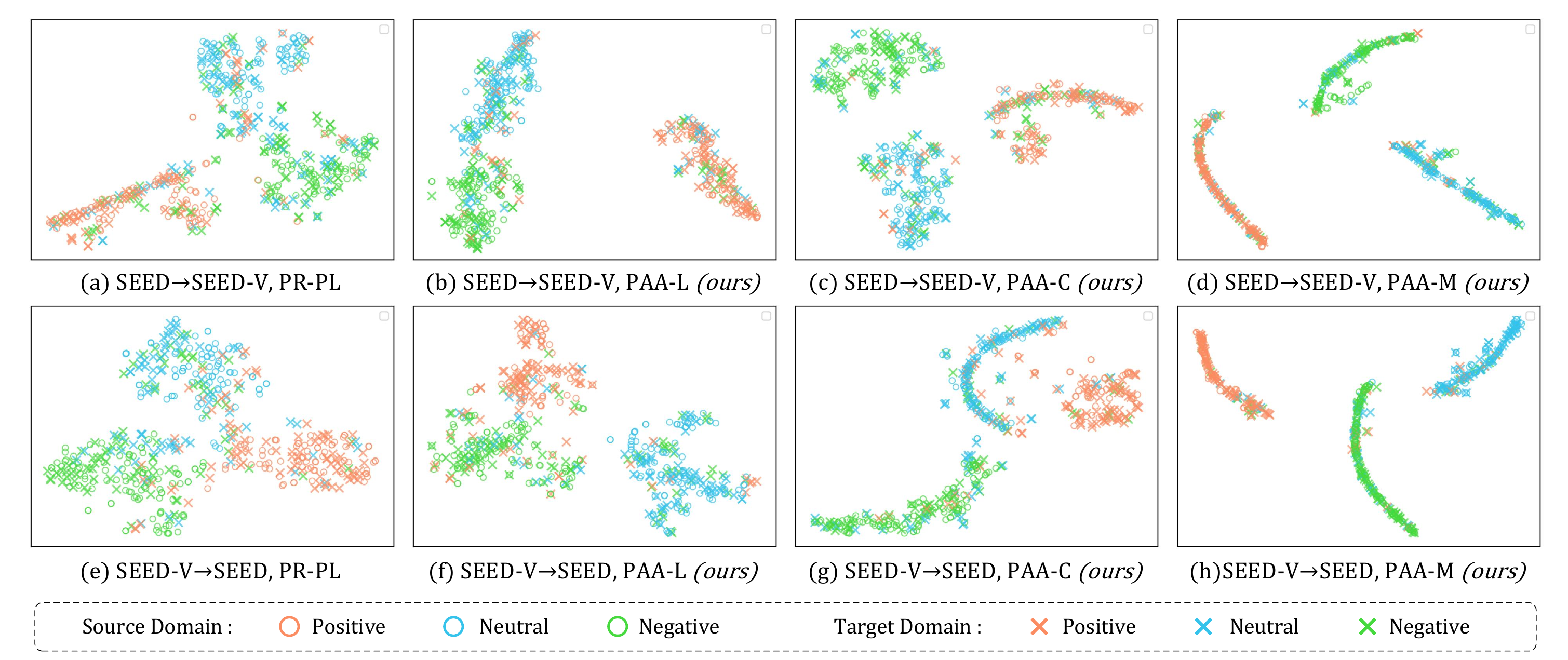}}
\caption{
    \textcolor{black}{T-SNE visualization of baseline model PR-PL and proposed PAA-L, PAA-C, and PAA-M.}
}
\label{fig:Vision}
\end{figure*}

\begin{table}[th]
\begin{center}
\caption{\textcolor{black}{The results of ablation experiment}}
\label{tab:Ablation_Experiment}
\scalebox{1}{
\color{black}
\begin{tabular}{lccc}
\toprule
\textit{Ablation Setting}    
    & PAA-L    & PAA-C     & PAA-M \\
\midrule
\textit{w/o RaL in $\mathcal{S}$ }
    & 55.21 ± 4.84     & 54.19 ± 6.42    & 65.11 ± 7.51 \\
\textit{w/o RaL in $\mathcal{S}$ and $\mathcal{T}$}
    & 50.48 ± 7.46	   & 47.39 ± 4.08	 & 63.66 ± 5.09 \\
\textit{w/o Prototype Repr.}
    & 54.46 ± 3.66	   & 52.88 ± 6.20	 & 58.62 ± 5.39 \\
\textit{w/o Trans. Matrix}
    & 58.43 ± 9.11	   & 62.50 ± 7.24	 & 66.72 ± 9.98 \\
\textit{w/o Feature Disc.}
    & 55.82 ± 5.30	   & 57.44 ± 5.23	 & 59.74 ± 10.6 \\
\midrule
\textit{w/o Stage (2)}
    & ---              & ---              & 63.25 ± 8.14 \\
\textit{w/o Stage (3)}
    & ---              & ---              & 58.42 ± 8.32 \\
\textit{w/o Stage (2)} \& \textit{(3)}
    & ---              & ---              & 58.38 ± 7.11 \\
\midrule
\textit{with all} 
    & \textbf{59.08 ± 7.83}	  
    & \textbf{63.74 ± 7.51}
    & \textbf{68.75 ± 6.03} \\
\bottomrule
\end{tabular}
}
\end{center}
\end{table}

\begin{figure}[th]
\centering
\subfloat{\includegraphics[width=0.45\textwidth]{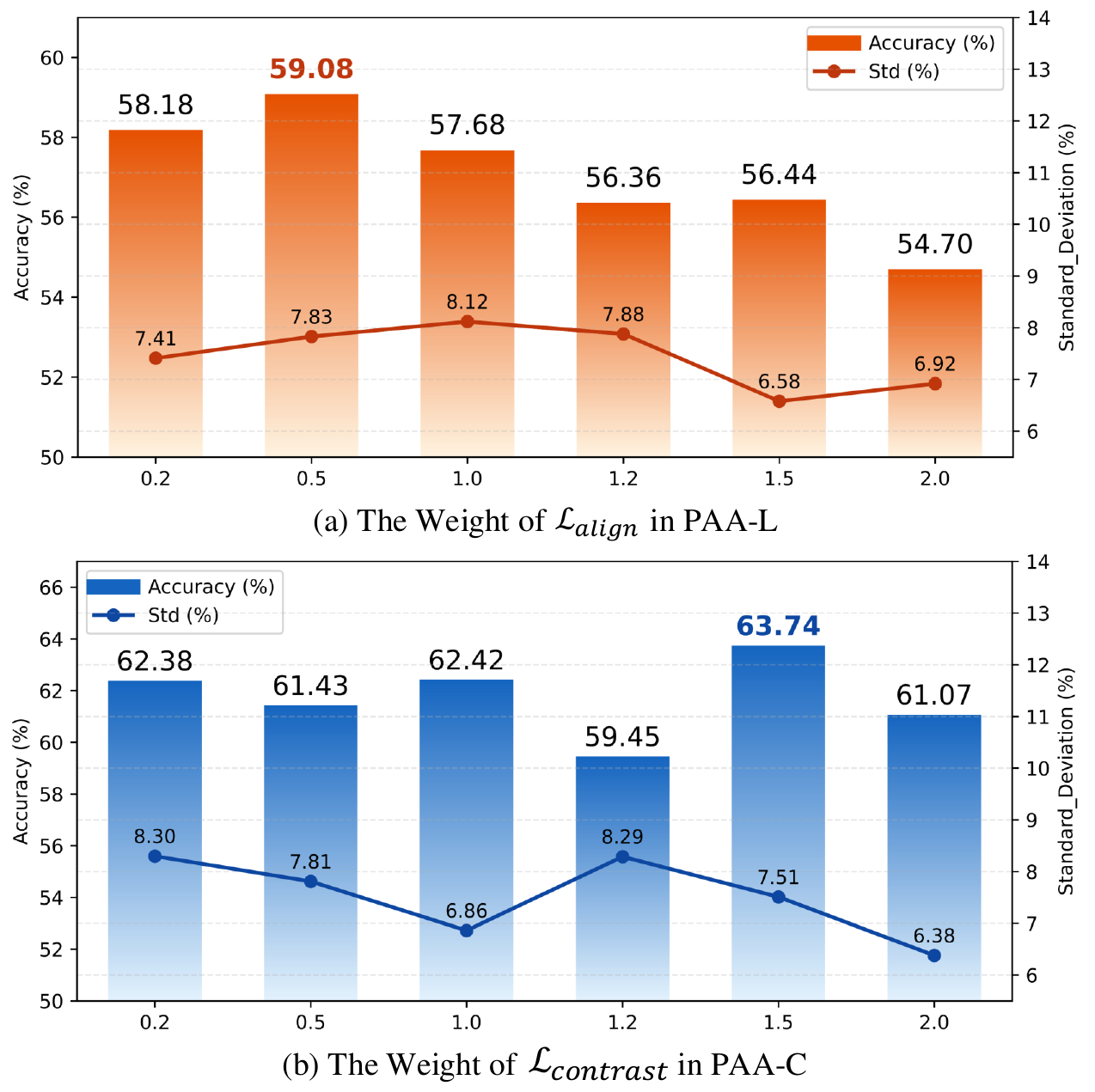}}
\caption{\textcolor{black}{Results for different parameter settings of the proposed PAA-L and PAA-C.}}
\label{fig:Parameter Sensitivity}
\end{figure}
\begin{figure}[th]
\centering
\subfloat{\includegraphics[width=0.49\textwidth]{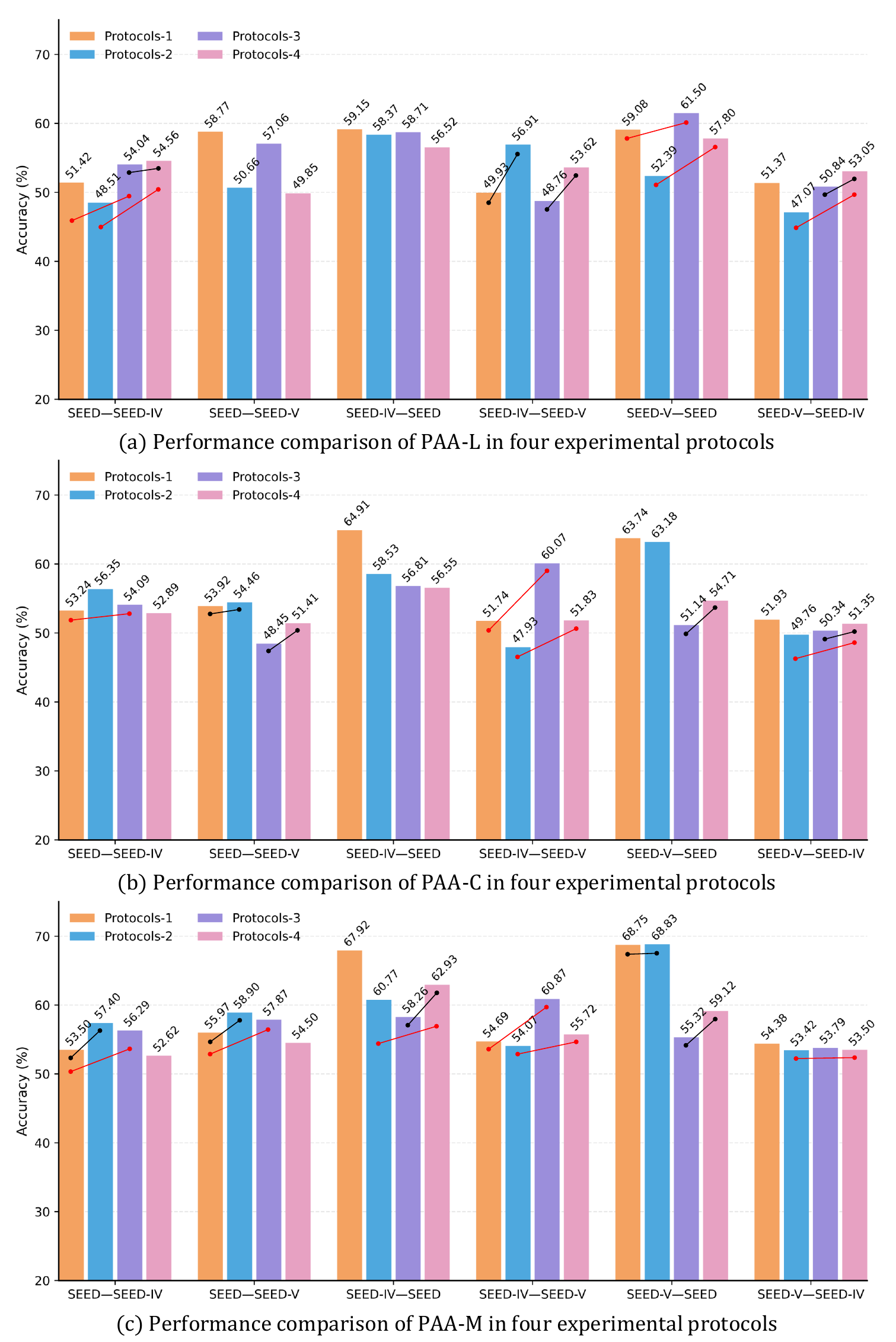}}
\caption{\textcolor{black}{Performance analysis of the proposed PAA-L, PAA-C, and PAA-M under four experimental protocols.}}
\label{fig:All_performances}
\end{figure}

\section{Discussion}
\label{Discussion}

\subsection{Confusion Matrix}
    \textcolor{black}{\indent To qualitatively evaluate the recognition performance of different optimization strategies for different emotion categories, we visualize the confusion matrix. The results of SEED$\to$SEED-V are shown in Fig.~\ref{fig:confusion_matrix} (a)$\sim$(d), where the accuracy of PAA-M in correctly recognizing negative, neutral, and positive emotions is 55.33\%, 52.25\%, and 72.53\%, respectively, which achieves the best performance among all methods. In negative and positive emotions, the PAA-C achieves suboptimal recognition performance with accuracies of 52.42\% and 56.52\%, respectively. In neutral emotion, the PR-PL model achieves suboptimal performance with an accuracy of 51.63\%.
    The results of SEED-V$\to$SEED are shown in Fig.~\ref{fig:confusion_matrix} (e)$\sim$(h), the PAA-M achieves the best performance in recognizing neutral and positive emotions, achieving 66.72\% and 79.91\% accuracy, respectively. The accuracy of PAA-L model for negative emotion is 68.11\%, while the accuracy of PAA-M is 66.34\%, the latter achieves suboptimal performance. Overall, all three strategies outperform the baseline model. PAA-M still shows the optimal performance in cross-corpus emotion recognition, while PAA-L and PAA-C are slightly better than PR-PL models.}

\subsection{Noise Robustness}
    \textcolor{black}{\indent In all the proposed optimization strategies, we adopt the RaL module to suppress the influence of label noise. Different from the traditional Single-sample learning (SSL), RaL transforms the single-shot classification problem into the similarity problem between samples. To verify the effectiveness of RaL module, we generate a large number of random noisy labels and replace a part of ture labels in the source domain with noisy labels. We consider the noise ratio (NR) = 10\%, 20\%, 30\%, 40\%.
    The validation results are shown in TABLE~\ref{tab:noisy_labels_1}, and the SaL strategy effectively alleviates the impact of label noise. When the NR is from 10\% to 40\%, the average accuracy decreases from 57.04\% to 53.91\%, which is only reduced by 3.13\%. When the NR is from 10\% to 40\%, the average accuracy drops from 50.23\% to 44.57\%, whcih decrease of 5.61\%. In addition, as shown in TABLE~\ref{tab:noisy_labels_2}, without adding label noise (0\%), the accuracy of the two strategies is 58.90\% and 53.74\%, respectively, with a gap of 5.16\%. However, when the NR = 40\%, the accuracy are 54.22\% and 44.57\%, respectively, with a gap of 9.29\%. Overall, the proposed RaL module has a lower sensitivity to noise labels.}
    
\subsection{Representation Visualization}
    \textcolor{black}{\indent To provide an intuitive analysis of the learned representations by the baseline model and proposed model (PAA-L, -C, and -M), we use t-SNE to visualize the feature distribution. The feature distribution of SEED$\to$SEED-V is shown in Fig.~\ref{fig:Vision} (a)$\sim$(d), and SEED-V$\to$SEED shown in (e)$\sim$(h). The distribution of emotion feature representation of PR-PL model is relatively scattered, and the decision boundary is fuzzy, which indicates that the discriminative feature extraction ability of emotion is limited. In contrast, the feature distributions of PAA-L and PAA-C strategies exhibit increasingly strong feature clustering performance, with relatively clear decision boundaries between different emotions, reflecting better feature alignment potential. Significantly, the learned feature representation of the PAA-M model is the most compact and the decision boundary is clear, which demonstrates the excellent cross-corpus feature alignment capability of the PAA-M model. Overall, all these results demonstrate the progressive optimization process of the three instantiated models.}

\subsection{Ablation Experiment}
    \textcolor{black}{\indent To assess the contribution of each component in the proposed model, we conduct ablation experiments under the SEED-V$\to$SEED in Experiment Protocol-(1).
    The results are shown in TABLE~\ref{tab:Ablation_Experiment}. When we remove the \textit{RaL module on source domain}, the accuracies of PAA-L, -C, and -M are 55.21\%, 54.19\%, and 65.11\%, which have decreased by 3.87\%, 9.55\%, and 3.64\% respectively. Similarly, when the \textit{RaL module on source and target domain} are removed, the accuracy of all methods also decreases significantly, to 50.48\%, 47.39\%, 63.66\% respectively. These results indicate that the RaL methods effectively enhances the performance and effectiveness of the model.
    The performance of all methods decreases when the \textit{Prototype Representation} is removed, which indicates that prototype representation learning helps to enhance the robustness of the model. 
    We introduced a trainable bilinear transformation matrix $\theta$ in feature interaction, which can enhance the robustness and generalization ability of the model. Removing the \textit{Transformation Matrix} results in 0.65\%, 1.24\%, and 2.03\% performance degradation for PAA-L, -C, and -M respectively. 
    When the \textit{Featre Discriminator} is removed, the accuracy of all methods are 55.82\%, 57.44\% and 59.74\%, and the performance decreases by 3.26\%, 6.30\% and 9.01\%, respectively. These results show the great potential of adversarial learning for feature alignment in cross domains.
    The training process of PPA-M consists of three stages. When removed the \textit{Stage (2)}, the model performance dropped from 68.75\% to 63.25\%. When we removed \textit{Stage (3)}, the model performance dropped from 68.75\% to 58.24\%. When both are removed, the model performance drops significantly. This further proves the significance of the three stage training process in PAA-M. 
    Overall, the ablation results show that each component contributes to the final performance of cross-corpus emotion recognition.}

\subsection{Parameter Sensitivity}
    \textcolor{black}{\indent Further, we quantitatively evaluate the sensitivity and effectiveness of the proposed PAA-L and PAA-C under different hyperparameter weights. In Eq.\ref{Eq:total}, the optimization objective of PAA-L has an important hyperparameter: the weight $\lambda_1$ of the class-conditional alignment $\mathcal{L}_{align}$. The significant component of the PAA-C optimization objective is the weight $\lambda_2$ of contrastive semantic discrepancy $\mathcal{L}_{contrast}$. They are both core parts of the  unified adversarial optimization objective and significantly affect the performance of the model. We conducted the test in the Experimental Protocol-1, the results are shown in the Fig.~\ref{fig:Parameter Sensitivity}. When $\lambda_1 = 0.5$, PAA-L achieves the best performance with an accuracy of 59.08\%. When $\lambda_2=1.5$, PAA-C achieves the Optimal performance, and the accuracy is 63.74\%.}

\subsection{Performance Analysis}
    \textcolor{black}{\indent All the performances of proposed models (PAA-L, -C, and -M) under the four experiment protocols are shown in Fig.~\ref{fig:All_performances} (a)$\sim$(c).
    Among them, Protocols-1 and Protocols-3 only adopt single-session data, while Protocols-2 and Protocols-4 adopt all sessions. The experimental protocol of the latter is stricter, but the performance does not always decline under the strict protocol (red mark, such as \textit{(a) SEED-IV$\to$SEED-V} and \textit{(c) SEED-V$\to$SEED}). This phenomenon might be due to the trade-off between data diversity and distribution pairing, which determines the final generalization performance. Increasing the amount of data may have amplified the benefits of data diversity under cross-corpus conditions. In addition, using multiple sessions during training may lead to implicit regularization mechanisms, which improve generalization cross corpus.
    \\ \indent The Protocols-3 and Protocols-4 are more stricter than Protocols-1 and Protocols-2 due to the use of independent unseen target data, and has greater challenges to the application potential of the model. Similarly, stricter protocols don't always lead to performance degradation (black mark, such as \textit{(a) SEED$\to$SEED-IV} and \textit{(b) SEED-IV$\to$SEED-V}). This phenomenon is because of the trade-off between training data utilization and generalization robustness, the independent test set may lead to more stable model selection. In addition, EEG signals are non-stationary and noise sensitive, and the impact of different protocols is inherently data dependent, which leads to non-monotonic performance changes in cross-corpus training. 
    \\ \indent In addition, databases with different experimental paradigms are also one of the reasons for the above two phenomena, which may also bring potential non-monotonic performance changes. Overall, although our proposed method achieves competitive results, there remains significant room for improvement in the field of cross-corpus emotion recognition.}

\begin{table}[]
\begin{center}
\caption{\textcolor{black}{The identification results of the Major Depressive Disorder. Here, the results of the baseline model reproduced by us are indicated by '*'.}}
\label{tab:MDD}
\scalebox{0.99}{
\color{black}
\begin{tabular}{lccc}
\toprule
\textit{Methods} 
        & \textit{\scriptsize SEED$\to$MDD}   
        & \textit{\scriptsize SEED-IV$\to$MDD}  
        & \textit{\scriptsize SEED-V$\to$MDD} \\
\midrule
KNN*\cite{KNN}    
        & 52.87 ± 4.87	& 52.75 ± 5.00	& 58.48 ± 4.09 \\
RF*\cite{breiman2001random} 
        & 60.41 ± 4.42	& 48.75 ± 4.98	& 59.25 ± 5.62 \\
KPCA*\cite{KPCA1999}
        & 53.47 ± 5.70	& 49.30 ± 4.97	& 57.61 ± 4.07 \\
SA*\cite{SA2013}  
        & 47.40 ± 5.00	& 48.76 ± 5.99	& 53.20 ± 4.97 \\
TCA*\cite{TCA2010}  
        & 50.78 ± 4.84	& 49.19 ± 4.28	& 52.56 ± 4.92 \\
DDC*\cite{2014Deep}
        & 56.46 ± 4.71	& 55.85 ± 5.67	& \underline{62.75} ± 3.01 \\
DDA*\cite{DDA_2022}    
        & 57.13 ± 4.49	& 54.51 ± 4.93	& 61.84 ± 3.10 \\
DMMR*\cite{wang2024dmmr}
        & 56.95 ± 4.64	& 51.01 ± 4.99	& 60.00 ± 4.54 \\
PR-PL*\cite{Li_Liang_2023}    
        & \underline{62.67} ± 5.07	
        & \underline{59.56} ± 6.26	
        & 62.59 ± 4.51 \\
\midrule
\multirow{2}{*}{{\small‌ PAA-L} (\textit{ours})} 
        & 64.91 ± 5.26	& 64.87 ± 6.36	& \textbf{67.04 ± 3.30} \\
        & \textcolor{orange}{\scriptsize(+2.24)}      
        & \textcolor{orange}{\scriptsize(+5.31)}
        & \textcolor{orange}{\scriptsize(+4.29)}  \\
\cmidrule(lr){2-4}
\multirow{2}{*}{{\small‌ PAA-C} (\textit{ours})} 
        & 65.75 ± 4.93	& 62.05 ± 6.33	& 64.62 ± 5.05 \\
        & \textcolor{orange}{\scriptsize(+3.08)}      
        & \textcolor{orange}{\scriptsize(+2.49)}
        & \textcolor{orange}{\scriptsize(+2.42)}  \\
\cmidrule(lr){2-4}
\multirow{2}{*}{{\small‌ PAA-M} (\textit{ours})} 
        & \textbf{67.37 ± 6.17}	
        & \textbf{66.05 ± 6.53}	
        & 66.19 ± 4.50 \\
        & \textcolor{orange}{\scriptsize(+4.70)}      
        & \textcolor{orange}{\scriptsize(+6.49)}
        & \textcolor{orange}{\scriptsize(+3.44)}  \\
\bottomrule
\end{tabular}
}
\end{center}
\end{table}


\subsection{Mental Illness Recognition}
    \textcolor{black}{\indent Major depressive disorder (MDD) is a serious mental illness. Research have shown that MDD is closely associated with dysfunction of the emotion regulation system, manifested as persistent negative affect, which constitutes one of the core diagnostic criteria for MDD \cite{snyder2026cont}. Therefore, we adopted a clinical EEG-based MDD database to further verify the feasibility of the proposed model in the real world. This database was obtained from the Hospital Universitiy Sains Malaysia, including 27 healthy subjects and 29 subjects diagnosed with MDD \cite{mumtaz2015review}. We consider the emotion database as the source domain and the MDD database as the target domain. There won't be any data leakage. The results are shown in TABLE~\ref{tab:MDD}, PAA-M achieves the SOTA performance in SEED$\to$MDD and SEED-IV$\to$MDD, which are 67.37\% and 66.05\%, respectively, an improvement of 4.7\% and 6.49\%. PAA-L achieves a performance of 67.04 with SEED-V$\to$MDD. These results further demonstrate the significant correlation between negative emotions and MDD, and also demonstrate the potential application of PAA in MDD.}

\section{Conclusion}
    \textcolor{black}{\indent This work propose a unified Prototype-driven Adversarial Alignment (PAA) framework for EEG emotion recognition. Three cross-corpus instantiation scenarios are implemented based on PPA: PAA-L performs local class-conditional alignment; PPA-C implements contrastive semantic regularization; PPA-M achieves complete boundary awareness. And the model is extended to clinical MDD application. Experiments show that the proposed method outperforms the baseline model. This work provides a potential solution for cross-corpus emotion recognition.}

\section*{Acknowledgements}
\textcolor{black}{\indent This work was supported in part by the National Natural Science Foundation of China under Grant 62176089, 62276169 and 62201356, in part by the Natural Science Foundation of Hunan Province under Grant 2023JJ20024, in part by the Key Research and Development Project of Hunan Province under Grant 2025QK3008, in part by the Key Project of Xiangjiang Laboratory under Granted 23XJ02006, in part by the STI 2030-Major Projects 2021ZD0200500, in part by the Medical-Engineering Interdisciplinary Research Foundation of Shenzhen University under Grant 2024YG008, in part by the Shenzhen University-Lingnan University Joint Research Programme, and in part by Shenzhen-Hong Kong Institute of Brain Science-Shenzhen Fundamental Research Institutions (2023SHIBS0003).}


%




\ifCLASSOPTIONcaptionsoff
  \newpage
\fi

\bibliographystyle{IEEEtran}
\bibliography{references}

\end{document}